\documentclass{article}

\usepackage[numbers,sort]{natbib} 
\usepackage[preprint]{neurips_2023}

\usepackage[utf8]{inputenc} % allow utf-8 input
\usepackage[T1]{fontenc}    % use 8-bit T1 fonts
\usepackage{hyperref}       % hyperlinks
\usepackage{url}            % simple URL typesetting
\usepackage{booktabs}       % professional-quality tables
\usepackage{amsfonts}       % blackboard math symbols
\usepackage{nicefrac}       % compact symbols for 1/2, etc.
\usepackage{microtype}      % microtypography
\usepackage{xcolor}         % colors
\usepackage{graphicx}
\usepackage{dsfont}
\usepackage{caption}
\usepackage{subcaption}
\usepackage{graphics}
\usepackage{multirow, makecell}
\usepackage{wrapfig}
\usepackage{bm}
\usepackage{amsmath}
\usepackage{enumitem}
\usepackage{algorithm}
\usepackage{algorithmic}
\usepackage{bbm}

\usepackage[capitalize,noabbrev]{cleveref}

\usepackage{amsmath}
\usepackage{amssymb}
\usepackage{mathtools}
\usepackage{amsthm}

\def\AUROC{\texttt{ROC}}
\def\ROC{\texttt{ROC}}
\def\TNR{\texttt{TNR}}

\usepackage[capitalize,noabbrev]{cleveref}

\theoremstyle{plain}

\theoremstyle{definition}

\theoremstyle{remark}

\usepackage[textsize=tiny]{todonotes}

\title{A Functional Data Perspective and Baseline On\\Multi-Layer Out-of-Distribution Detection}

\author{%
Eduardo Dadalto$^{1,2,3,4}$ \quad Pierre Colombo$^{2,5}$ \quad Guillaume Staerman$^{2,6,7}$\\\textbf{Nathan Noiry}$^8$ \quad \quad \textbf{Pablo Piantanida}$^{2,3,4,9}$\\
$^1$Laboratoire des signaux et systèmes (L2S)\\
$^2$Université Paris-Saclay \quad $^3$CNRS \quad $^4$CentraleSupélec\\
$^5$MICS \quad $^6$Inria \quad $^7$CEA \quad $^8$Owkin\\
$^9$International Laboratory on Learning Systems (ILLS)\\
\texttt{\{eduardo.dadalto,pierre.colombo\}@centralesupelec.fr}\\
\texttt{guillaume.staerman@inria.fr} \quad \texttt{noirynathan@gmail.com}\\
\texttt{pablo.piantanida@cnrs.fr}
}

\begin{document}

\maketitle

\begin{abstract}
% A crucial component for implementing reliable classifiers is detecting examples far from the reference (training) distribution, referred to as out-of-distribution (OOD) samples. 
A key feature of out-of-distribution (OOD) detection is to exploit a trained neural network by extracting statistical patterns and relationships through the multi-layer classifier to detect shifts in the expected input data distribution.
Despite achieving solid results, several state-of-the-art methods rely on the penultimate or last layer outputs only, leaving behind valuable information for OOD detection. Methods that explore the multiple layers either require a special architecture or a supervised objective to do so.
This work adopts an original approach based on a functional view of the network that exploits the sample's trajectories through the various layers and their statistical dependencies. It goes beyond multivariate features aggregation and introduces a baseline rooted in functional anomaly detection. In this new framework, OOD detection translates into detecting samples whose trajectories differ from the typical behavior characterized by the training set. 
We validate our method and empirically demonstrate its effectiveness in OOD detection compared to strong state-of-the-art baselines on computer vision benchmarks\footnote{Our code is available online at \url{https://github.com/edadaltocg/ood-trajectory-projection}.}.
\end{abstract}

\section{Introduction}\label{sec:introduction}

The ability of a Deep Neural Network (DNN) to generalize to new data is mainly restricted to priorly known concepts in the training dataset. In real-world scenarios,  Machine Learning (ML) models may encounter Out-Of-Distribution (OOD) samples, such as data belonging to novel concepts (classes)~\citep{pimentel2014review}, abnormal samples \citep{tishby2015deep}, or even carefully crafted attacks designed to exploit the model \citep{szegedy2013intriguing}. The behavior of ML systems on unseen data is of great concern for safety-critical applications \citep{problems-in-ai-safety,amodei2016concrete}, such as medical diagnosis in healthcare \citep{adarsh_health_ai_2020}, autonomous vehicle control in transportation \citep{bojarski2016end}, among others. To address safety issues arising from OOD samples, a successful line of work aims to augment ML models with an OOD binary detector to distinguish between abnormal and in-distribution examples~\citep{baseline}. An analogy to the detector is the human body's immune system, with the task of differentiating between antigens and the body itself. 

Distinguishing OOD samples is challenging. Some previous works developed detectors by combining scores at the various layers of the multi-layer pre-trained  classifier~\citep{gram_matrice,mahalanobis,gomes2022igeood,Huang2021OnTI_gradnorm,colombobeyond}. These detectors require either a held-out OOD dataset (e.g., adversarially generated data) or ad-hoc methods to combine OOD scores computed on each layer embedding. A key observation is that existing aggregation techniques overlook the sequential nature of the underlying problem and, thus, limit the discriminative power of those methods. Indeed, an input sample passes consecutively through each layer and generates a highly correlated signature that can be statistically characterized. Our observations in this work motivate the statement:
\begin{center}
\textit{The input's trajectory through a network is key for distinguishing typical samples from atypical ones.}
\end{center}

In this paper, we introduce a significant change of perspective. Instead of looking at each layer score independently, we cast the scores into a sequential representation that captures the statistical trajectory of an input sample through the various layers of a multi-layer neural network. To this end, we adopt a functional point of view by considering the sequential representation as curves parameterized by each layer. Consequently, we redefine OOD detection as detecting samples whose trajectories are abnormal (or atypical) compared to reference trajectories characterized by the training set. Through a vast experimental benchmark, we showed that the functional representation of a sample encodes valuable information for OOD detection.

\textbf{Contributions.} This work brings novel insights into the problem of OOD detection. It presents a method for detecting OOD samples without hyperparameter tuning and no additional outlier data. Our main contributions are summarized as follows.
\begin{enumerate}%[wide, topsep=0pt, labelwidth=0pt, labelindent=0pt]
    % \item \textit{A novel problem formulation}. We reformulate the problem of OOD detection through a functional perspective that effectively captures the statistical dependencies of an input sample's path across a multi-layer neural classifier. Moreover, we propose a map from the multivariate feature space (at each layer) to a functional space that relies on the probability weighted projection of the test sample onto the class conditional training prototypes at the layer. 
    %It is computationally efficient and straightforward to implement.
    \item \textit{Computing OOD scores from trajectories}. We propose a semantically informed map from multiple embedding spaces to piecewise linear functions. Subsequently, the simple inner product between the test sample's trajectory and the training prototype trajectory indicates how likely a sample is to belong to in-distribution.
    % Low similarity indicates that the test sample is likely sampled from OOD.
    \item \textit{Extensive empirical evaluation}. We validate the value of the proposed method by demonstrating gains against twelve strong state-of-the-art methods on both CIFAR-10 and ImageNet on average TNR at 95\% TPR and AUROC across five NN architectures.
    %We release our code \href{https://github.com/ood-trajectory/ood-trajectory}{online}.
\end{enumerate}

\section{Related Works}\label{sec:related_work}

This section briefly discusses prior work in OOD detection, highlighting confidence-based and feature-based methods without special training as they resonate the most with our work. Another thread of research relies on learning representations adapted to OOD detection~\citep{Mohseni_Pitale_Yadawa_Wang_2020, bitterwolf2021certifiably, mahmood2021multiscale,du2022siren}, either through contrastive training \citep{hendrycks2019OE_ood, contrastive_training_ood,Sehwag2021SSDAU}, regularization \citep{lee2021removing, nandy2021maximizing, hein2019relu,Du2022VOSLW}, generative \citep{anoGAN, ood_via_generation,xiao2020likelihood,ren2019likelihoodratios,zhang2021outofdistribution}, or ensemble \citep{vyas2018outofdistribution,choi2019waic} based approaches. Related subfields are open set recognition \citep{Geng2021open_set_survey}, novelty detection \citep{pimentel2014review}, anomaly detection \citep{deep_anom_detec}, outlier detection \citep{Hodge2004_outlier_detection}, and adversarial attacks detection \citep{adv_attack_2018_survey}.

\textbf{Confidence-based OOD detection.}
A natural measure of uncertainty of a sample's label is the classification model's softmax output \cite{baseline}. 
% \cite{baseline} observed that the maximum of the softmax output could be used as a discriminative score between in-distribution and OOD samples.
\cite{hein2019relu} observed that it may still assign overconfident values to OOD examples. \cite{odin} and \cite{Hsu2020GeneralizedOD} propose re-scaling the softmax response with a temperature value and a pre-processing technique that further separates in- from out-of-distribution examples. \cite{liu2020energybased} proposes an energy-based OOD detection score by replacing the softmax confidence score with the free energy function. \cite{Hendrycks2022ScalingOD} computes the KL divergence between the test data probability vectors and the class conditional training logits  prototypes. \cite{sun2022dice} proposes sparsification of the classification layer weights to improve OOD detection by regularizing predictions. While \cite{Wang2022ViMOW} recomputes the logits with information coming from the feature space by projecting them in a new coordinate system and recomputing the logits.

\textbf{Feature-based OOD detection.} This line of research focuses on exploring latent representations for OOD detection. For instance, \cite{haroush2021statistical} considers using statistical tests; \cite{gram_matrice} rely on higher-order Grams matrices; \cite{quintanilha2019detecting} uses mean and standard deviation within feature maps; \cite{Sun2021ReActOD} proposes clipping the activations to boost OOD detection performance; while recent work \cite{Huang2021OnTI_gradnorm} also explores the gradient space and modifying batch normalization \citep{zhu2022boosting}. Normalizing and residual flows to estimate the probability distribution of the feature space were proposed in \cite{kirichenko2020normalizing,Zisselman2020DeepRF}. \cite{Dong2021NeuralMD} regards the average activations of intermediate features' and trains a lightweight classifier on top of them. \cite{Sun2022OutofdistributionDW} proposes a non-parametric nearest-neighbor based on the Euclidean distance for OOD detection. \cite{song2022rankfeat} removes the largest singular-vector from the representation matrices of two intermediate features and then computes the free energy over the mixed logits. Perhaps one of the most widely used techniques relies on the Gaussian mixture assumption for the hidden representations and the Mahalanobis distance \citep{mahalanobis, ren2021simple} or further information geometry tools \citep{gomes2022igeood}. 
%The reasons are its simplicity, performance, and geometrical explanation. 
Efforts toward combining multiple features to improve performance were previously explored in \citep{mahalanobis,gram_matrice,gomes2022igeood}. The strategy relies upon having additional data for tuning the detector or focusing on specific model architectures, which are limiting factors in real-world applications. For instance, MOOD~\cite{lin2021mood} relies on the MSDNet architecture, which trains multiple classifiers on the output of each layer in the feature extractor, and their objective is to select the most appropriate layer in inference time to reduce the computation cost. On the other hand, we study the trajectory of an input through the network. Unlike MOOD, our method applies to any current architecture of NN.

\section{Preliminaries}
We start by recalling the general setting of the OOD detection problem from a mathematical point of view (Section~\ref{sec:background}). Then, in Section~\ref{sec:motivation}, we motivate our method through a simple yet clarifying example showcasing the limitation of previous works and how we approach the problem.

\subsection{Background}\label{sec:background}

Let $(X, Y)$ be a random variable valued in a space $\mathcal{X} \times \mathcal{Y}$ with unknown probability density function (pdf) $p_{XY}$ and probability distribution $P_{XY}$. Here, $\mathcal{X} \subseteq \mathbb{R}^d$ represents the covariate space and $\mathcal{Y}=\{1,\ldots, C\}$ corresponds to the labels attached to elements from $\mathcal{X}$. The training dataset  $\mathcal{S}_N ~=\{(\boldsymbol{x}_i,y_i) \}_{i=1}^N $ is defined as independent and identically distributed (i.i.d) realizations of $P_{XY}$. 
From this formulation, detecting OOD samples boils down to building a binary rule $g:\mathcal{X} \rightarrow \{0,1\}$ through a soft scoring function $s : \mathcal{X} \rightarrow \mathbb{R} $ and a threshold $\gamma \in \mathbb{R}$. Namely, a new observation $\boldsymbol{x} \in \mathcal{X}$ is then considered as \textit{in-distribution}, i.e., generated by $P_{XY}$, when $g(\boldsymbol{x})=0$ and as OOD when $g(\boldsymbol{x})=1$. Finding this rule $g$ from $\mathcal{X}$ can become intractable when the dimension $d$ is large. Thus, previous work rely on a multi-layer pre-trained classifier $f_\theta : \mathcal{X} \rightarrow \mathcal{Y}$ defined as:
\begin{equation*}\label{eq:deep_net}
    f_\theta(\cdot)= h \circ f_L \circ f_{L-1} \circ \dots \circ f_1 (\cdot),
\end{equation*} with $L\geq 1$ layers, where $f_{\ell}: \mathbb{R}^{d_{\ell-1}} \rightarrow \mathbb{R}^{d_{\ell}}$ is the $\ell$-th
layer of the multi-layer neural classifier, $d_{\ell}$ denotes the dimension of the latent space induced by the $\ell$-th layer ($d_0 = d$), and $h$ indicates the classifier that outputs the logits. We also define  $\boldsymbol{z}_{\ell}=(f_{\ell} \circ \dots \circ f_1)(\boldsymbol{x})$ as the latent vectorial representation at the $\ell-$th layer for an input sample $\boldsymbol{x}$. We will refer to the logits as $\boldsymbol{z}_{L+1}$ and $h$ as $f_{L+1}$ to homogenize notation. It is worth emphasizing that the trajectory of $(\boldsymbol{z}_{1},\boldsymbol{z}_{2},\dots, \boldsymbol{z}_{L+1})$ corresponding to a test input $\boldsymbol{x}_0$ are dependent random variables whose joint distribution strongly depends on the underlying distribution of the input. 

Therefore, the design of function $g(\cdot)$ is typically based on the three key steps:
\begin{itemize}%[wide, topsep=0pt, labelwidth=0pt, labelindent=0pt]
    \item[(i)] A  similarity measure  ${\rm d}(\cdot\,;\cdot )$ (e.g., Cosine similarity, Mahalanobis distance, etc.) between a sample and a population is applied at each layer to measure the similarity (or dissimilarity) of a test input $\boldsymbol{x}_0$ at the $\ell$-th layer $\boldsymbol{z}_{\ell,0}=(f_{\ell} \circ \dots \circ f_1)(\boldsymbol{x}_0)$ w.r.t. the population of the training examples observed at the same layer  $\big\{\boldsymbol{z}_{\ell}=(f_{\ell} \circ \dots \circ f_1)(\boldsymbol{x}) \, :\, \boldsymbol{x}\in \mathcal{S}_N\big\}$. 
%     (e.g., Mahalanobis distance ${\rm d}_{\rm M}$)
    \item[(ii)] The layer-wise score obtained is mapped to the real line collecting the OOD scores.
    \item[(iii)] A threshold is set to build the final decision function.
\end{itemize}
A fundamental ingredient remains in step (ii): 

\begin{center}
\textit{How to consistently leverage the information collected from multiple layers outputs in an unsupervised way, i.e., without resorting to OOD or pseudo-OOD examples?}
\end{center}

\subsection{From Independent Multi-Layer Scores to a Sequential Perspective of OOD Detection}\label{sec:motivation}
\begin{figure*}[!htp]
     \centering
    \begin{subfigure}[b]{0.32\textwidth}
        \includegraphics[width=\textwidth]{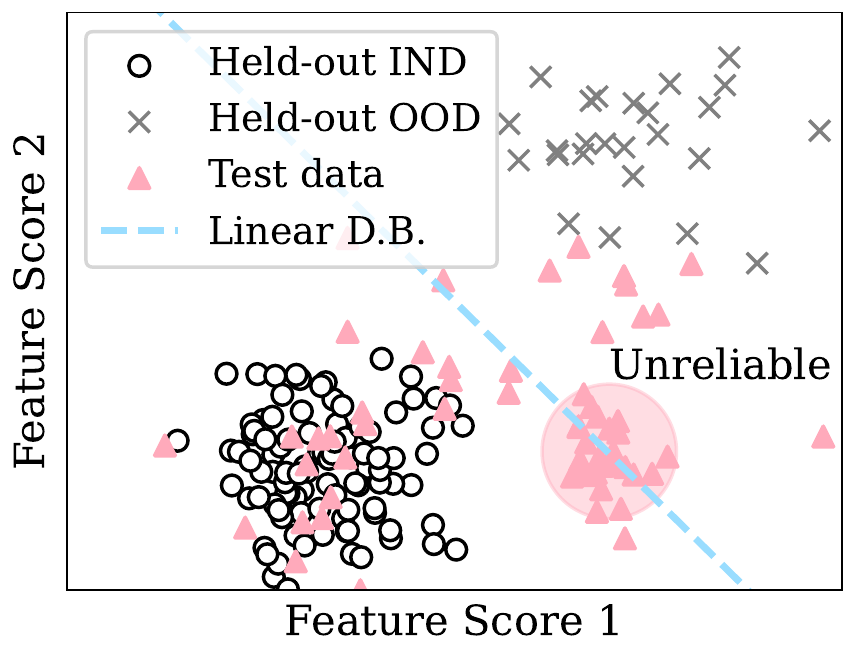}
        \caption{Example of a misspecified model in a toy example in 2D caused by fitting with held-out OOD dataset.}
        \label{fig:toy_example_2d}
    \end{subfigure}
    \hfill
    \begin{subfigure}[b]{0.32\textwidth}
        \centering
        \includegraphics[width=\textwidth]{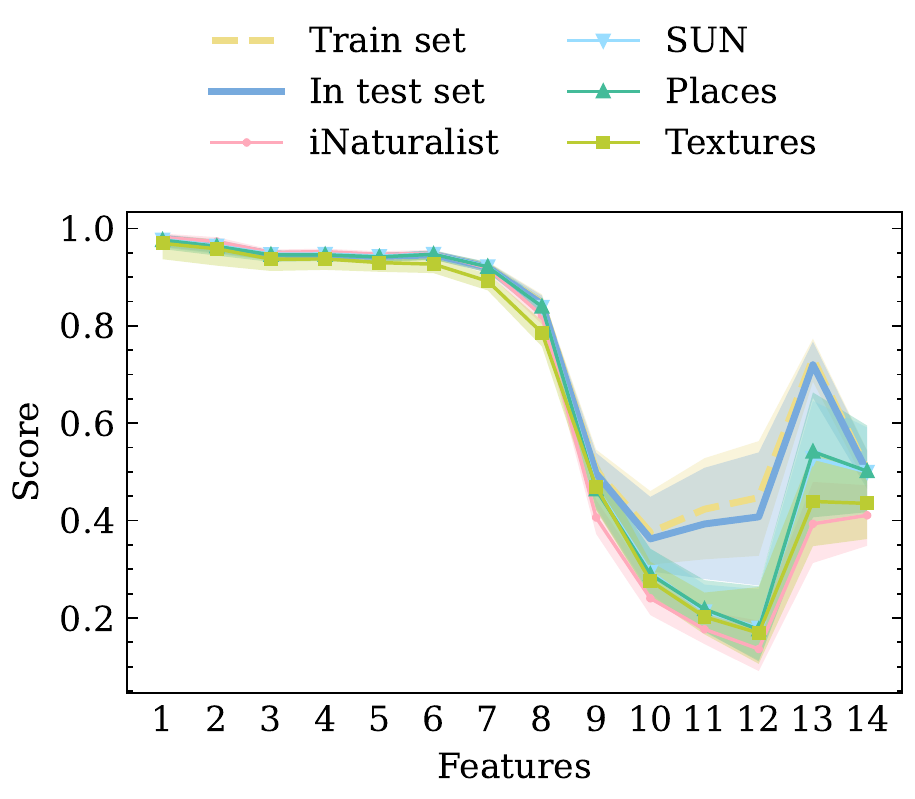}
        \caption{Trajectory of data through a network with 25\% and 75\% percentile bounds. }
        \label{fig:trajectories_example}
    \end{subfigure}
    \hfill
    \begin{subfigure}[b]{0.32\textwidth}
        \centering
        \includegraphics[width=\textwidth]{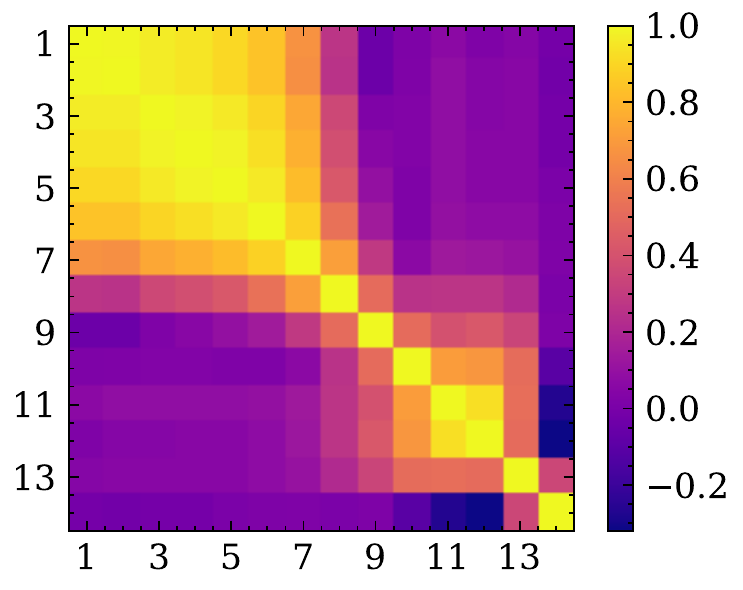}
        \caption{Correlation between layers training scores in a network, highlighting structure in the trajectories.}
        \label{fig:corr_mat}
    \end{subfigure}
    \caption{Figure \ref{fig:toy_example_2d} summarizes the limitation of supervised methods for aggregating layer scores that rely on held-out OOD or pseudo-OOD data. It biases the decision boundary (D.B) that does not generalize well to other types of OOD data. We observed that in-distribution and OOD data have disparate trajectories through a network (Fig. \ref{fig:trajectories_example}), especially on the last five features. These features are correlated in a sequential fashion, as observed in Fig. \ref{fig:corr_mat}.}
    \label{fig:time_hist_roc_main}
\end{figure*}
Previous multi-feature OOD detection works treat step (ii) as a supervised learning problem \citep{mahalanobis, gomes2022igeood} for which the solution is a linear binary classifier. The objective is to find a linear combination of the scores obtained at each layer that will sufficiently separate in-distribution from OOD samples. A held-out OOD dataset is collected from true  (or pseudo-generated) OOD samples. The linear soft novelty score functions $s_\alpha$ writes:
\begin{equation*}
    s_\alpha(\boldsymbol{x}_0) = \sum_{\ell=1}^{L} \alpha_\ell \cdot {\rm d} \left(\boldsymbol{x}_0 ; \big\{(f_{\ell} \circ \dots \circ f_1)(\boldsymbol{x}) \, :\, \boldsymbol{x}\in \mathcal{S}_N\big\} \right).
\end{equation*}
The shortcomings of this method are the need for extra data or ad-hoc parameters, which results in decision boundaries that underfit the problem and fail to capture certain types of OOD samples. To illustrate this phenomenon, we designed a toy example (see Figure~\ref{fig:toy_example_2d}) where scores are extracted from two features fitting a linear discriminator on held-out in-distribution (IND) and OOD samples. 

As a consequence, areas of unreliable predictions where OOD samples cannot be detected due to the misspecification of the linear model arise. One could simply introduce a non-linear discriminator that better captures the geometry of the data for this 2D toy example. However, it becomes challenging as we move to higher dimensions with limited data.

%\textbf{The sequential nature of feature extraction matters.} 
By reformulating the problem from a functional data point of view,
%\footnote{The method can look into the past and future of the series.}
we can identify trends and typicality in trajectories extracted by the network from the input. Figure~\ref{fig:trajectories_example} shows the dispersion of trajectories coming from the in-distribution and OOD samples. These patterns are extracted from multiple latent representations and aligned on a time-series-like object.  \textit{We observed that trajectories coming from OOD samples exhibit a  different shape when compared to typical trajectories from training data}. Thus, to determine if an instance belongs to in-distribution, we can test if the observed path is similar to the functional trajectory reference extracted from the training set. 

\section{Towards Functional Out-Of-Distribution Detection}
This section presents our OOD detection framework, which applies to any pre-trained multi-layer neural network with no requirements for OOD samples. We describe our method through two key steps: functional representation of the input sample (see Section~\ref{sec:3.1}) and test time OOD score computation (see Section~\ref{sec:3.2}).

\subsection{Functional Representation}\label{sec:3.1}

The first step to obtaining a univariate functional representation of the data from the multivariate hidden representations is to reduce each feature map to a scalar value. To do so, we first compute the class-conditional training population prototypes defined by:
\begin{equation}\label{eq:mean_cov}
    \boldsymbol{\mu}_{\ell,y} = \frac{1}{N_y}\sum_{i=1}^{N_y} \boldsymbol{z}_{\ell, i},
\end{equation}
where $N_y = \big|\{\boldsymbol{z}_{\ell, i} :   y_i = y, \forall i \in \{1..N\}\}\big|$, $1\leq \ell \leq L+1$ and $ \boldsymbol{z}_{\ell, i}=(f_{\ell} \circ \dots \circ f_1)(\boldsymbol{x}_i)$.

Given an input example, we compute the \textit{probability weighted scalar projection}\footnote{Other metrics to measure the similarity of an input w.r.t. the population of examples can also be used.} between its features (including the logits) and the training class conditional prototypes, resulting in $L+1$ scalar scores:
\begin{equation}\label{eq:metric}
{\rm d}_\ell(\boldsymbol{x}; \mathcal{M}_{\ell})=
\sum_{y=1}^{C}\sigma_y(\boldsymbol{x}) \cdot {\rm proj}_{ \boldsymbol{\mu}_{\ell, y}}\boldsymbol{z}_{\ell}  
 = \sum_{y=1}^{C}\sigma_y(\boldsymbol{x}) \rVert \boldsymbol{z}_{\ell} \lVert\cos\bigl(\angle\left( \boldsymbol{z}_{\ell}, \boldsymbol{\mu}_{\ell, y} \right)\bigr),
\end{equation}
where $\mathcal{M}_{\ell} = \{\boldsymbol{\mu}_{\ell,y} : y \in \mathcal{Y}\}$, $\lVert \cdot \rVert$ is the $\ell_2$-norm, $\angle\left(\cdot, \cdot\right)$ is the angle between two vectors, and $\sigma_y(\boldsymbol{x};f_\theta)$ is the softmax function on the logits $f_\theta(\boldsymbol{x})$ of class $y$.
Hence, our layer-wise scores rely on the notions of vector length and angle between vectors, which can be generalized to any $n$-dimensional inner product space without imposing any geometrical constraints.

It is worth emphasizing that our layer score has some advantages compared to the class conditional Gaussian model first introduced in~\cite{mahalanobis} and the Gram matrix based-method introduced in~\cite{gram_matrice}. Our layer score encompasses a broader class of distributions as we do not suppose a specific underlying probability distribution. We avoid computing covariance matrices, which are often ill-conditioned for latent representations of DNNs \cite{Ahuja2019ProbabilisticMO}. Since we do not store covariance matrices, our functional approach has a negligible overhead regarding memory requirements. Also, our method can be applied to any vector-based hidden representation, not being restricted to matrix-based representations as in~\cite{gram_matrice}. Thus, our approach applies to a broader range of models, including transformers.
%, in which the hidden representation is usually taken as the [CLS] latent vector token.

By computing the scalar projection at each layer, we define the following \textit{functional neural-representation} extraction function given by Eq. \ref{eq:time_series}. Thus, we can map sample representations to a functional space while retaining information on the typicality w.r.t the training dataset.
\begin{align}
    \begin{split}\label{eq:time_series}
        \phi: \mathcal{X} &\rightarrow \mathbb{R}^{L+1}\\
        \boldsymbol{x} &\mapsto \big[{\rm d}_{1} \left(\boldsymbol{x}; \mathcal{M}_{1} \right), \dots, {\rm d}_{L+1} \left(\boldsymbol{x}; \mathcal{M}_{L+1} \right) \big]
    \end{split}
\end{align}
We apply $\phi$ to the training input $\boldsymbol{x}_i$ to obtain the  representation of the training sample across the network  $\boldsymbol{u}_i = \phi(\boldsymbol{x}_i)$. We consider the related vectors $\boldsymbol{u}_i, \forall~i\in [1:N]$\footnote{We observed empirically that subsampling vectors to even $N$/$100$ yields very good results.} as curves parameterized by the layers of the network.  We build a training reference dataset $\mathcal{U} = \{\boldsymbol{u}_i\}_{i=1}^{N}$ from these functional representations that will be useful for detecting OOD samples during test time.
% \begin{equation}
%     \mathcal{U} = \{\boldsymbol{u}_i\}_{i=1}^{N}.
% \end{equation}
We then rescale the training set trajectories w.r.t the maximum value found at each coordinate to obtain layer-wise scores on the same scaling for each coordinate. Hence, for $j\in \{1,\dots,L+1\}$, let $\max\left({\mathcal{U}}\right) := [\max_i {\boldsymbol{u}_{i,1}}, \dots, \max_i \boldsymbol{u}_{i,L+1}]^\top$, we can compute a reference trajectory $\bar{\boldsymbol{u}}$ for the entire training dataset defined in \eqref{eq:typical_traj} that will serve as a global \textit{typical reference} to test trajectories.
\begin{equation}\label{eq:typical_traj}
    \bar{\boldsymbol{u}} = \frac{1}{N } \sum_{i=1}^N \frac{\boldsymbol{u}_i}{\max (\mathcal{U})}
\end{equation}

\subsection{Computing the OOD Score at Test Time}\label{sec:3.2}

\begin{figure*}[!htp]
    \centering
        \includegraphics[width=0.95\textwidth]{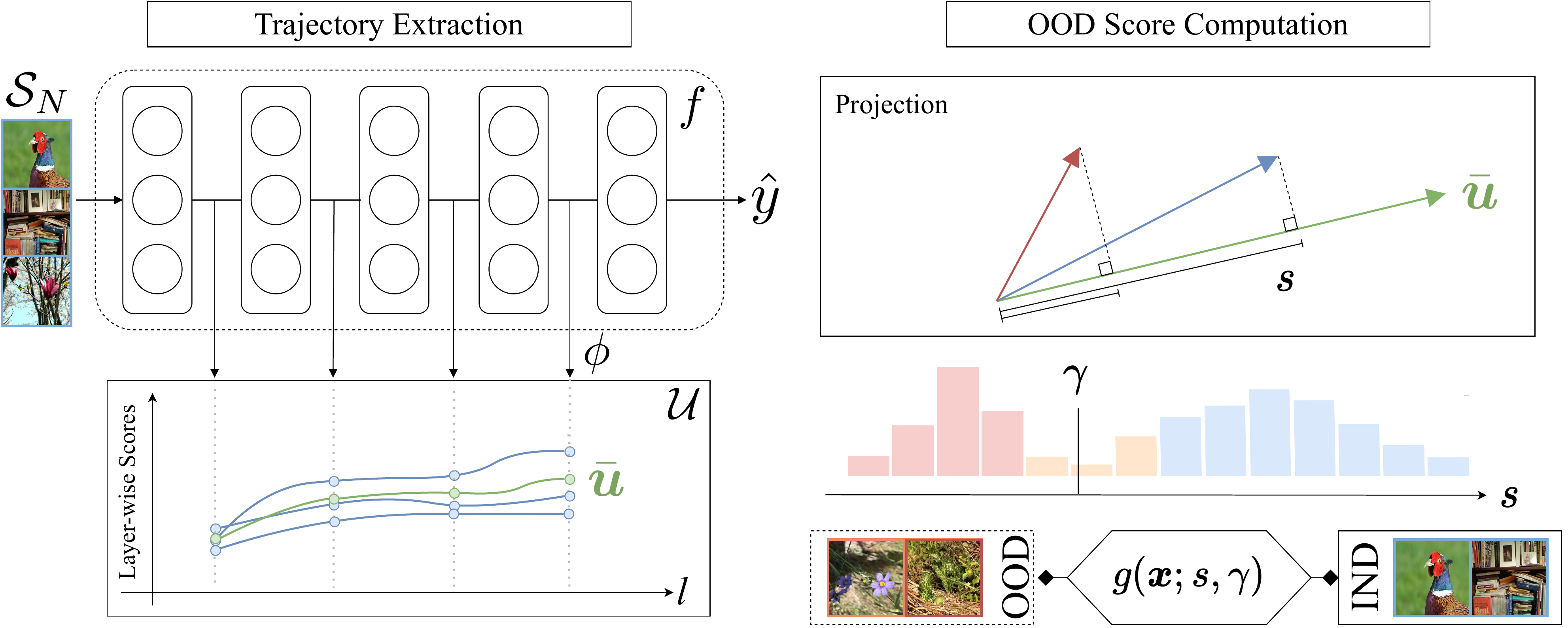}
      \caption{The left-hand side of the figure shows the feature extraction process of a deep neural network classifier $f$. The mapping of the hidden representations of an input sample into a functional representation is given by a function $\phi$. 
      The right-hand side of the figure shows how our method computes the OOD score $s$ of a sample during test time. The sample's trajectory is projected to the training reference trajectory $\bar{u}$. Finally, a threshold $\gamma$ is set to obtain a discriminator $g$.}
    \label{fig:main}
\end{figure*}

At inference time, we first re-scale the test sample's trajectory as we did with the training reference $ \bar{\phi}(\boldsymbol{x}) = \phi(\boldsymbol{x}) /\max\left(\mathcal{U}\right).$
Then, we compute a similarity score w.r.t this typical reference, resulting in our OOD score. We choose as metric also the scalar projection of the test vector to the training reference. In practical terms, it boils down to the \textit{inner product} between the test sample's trajectory and the training set's typical reference trajectory since the norm of the average trajectory is constant for all test samples. Mathematically, our scoring function $s:\mathcal{X}\mapsto\mathbb{R}$ writes:
\begin{equation}\label{eq:inner}
    s(\boldsymbol{x};\bar{\boldsymbol{u}}) = \bigl\langle\bar{\phi}(\boldsymbol{x}), 
        \bar{\boldsymbol{u}}\bigl\rangle
        = \sum_{j=1}^{L+1}\bar{\phi}(\boldsymbol{x})_j \bar{\boldsymbol{u}}_j
\end{equation}
which is bounded by Cauchy-Schwartz's inequality. From this OOD score, we can derive a binary classifier $g$ by fixing a threshold $\gamma\in \mathbb{R}$, $ g(\boldsymbol{x}; s, \gamma) = \mathbbm{1}_{ s(\boldsymbol{x}) \leq \gamma}$,
% \begin{equation}
%     g(\boldsymbol{x}; s, \gamma) = {\begin{cases}
%          & 1, \text{ if }s(\boldsymbol{x}) \leq \gamma\\
%          & 0, \text{ otherwise,}
%     \end{cases}}
% \end{equation}
where $g(\boldsymbol{x})=1$ means that the input sample $\boldsymbol{x}$ is classified as being out-of-distribution. Please refer to  Appendix (see Section~\ref{sec:alg_appendix}) for further details on the algorithm and \cref{fig:main} for an illustrated summary.

\textbf{\emph{Remark.}} Our proposed baseline is equivalent to a multivariate linear model over the vectorial trajectories, which can be viewed as a special case of the more general functional linear model \cite{ramsay2004functional}. For piecewise linear functions, the inner product in Euclidean space is equivalent to that in Hilbert space $L^{2}$. Thus, our theoretical framework can be viewed as an extension of the traditional multivariate models, expanding the OOD detection horizon towards solutions that explore more general Hilbertian spaces.

\section{Experimental Setting}\label{sec:experimental_setting}

% This section describes the experimental setting, including the datasets used, the pre-trained DNN architectures, the evaluation metrics, and other essential information.
% We open-source our code in \url{https://github.com/ood-trajectory/ood-trajectory}.

\textbf{Datasets.}
We set as \textit{in-distribution} dataset \textit{ImageNet-1K} (= ILSVRC2012; \citealp{imagenet}) for our main experiments, which is a challenging mid-size and realistic dataset. It contains around 1.28M training samples and 50,000 test samples belonging to 1000 different classes.
For the \textit{out-of-distribution} datasets, we take the same dataset splits introduced by \cite{Huang2021MOSTS}. The \textit{iNaturalist} \citep{Horn2017TheICINaturalist} split with 10,000 test samples with concepts from 110 classes different from the in-distribution ones. The \textit{Sun} \citep{sun} dataset with a split with 10,000 randomly sampled test examples belonging to 50 categories. The \textit{Places365} \citep{zhou2017places} dataset with 10,000 samples from 50 disjoint categories. For the DTD or \textit{Textures} \citep{cimpoi14describing} dataset, we considered all of the 5,640 available test samples. Note that there are a few overlaps between the semantics of classes from this dataset and ImageNet. We decided to keep the entire dataset in order to be comparable with \cite{Huang2021OnTI_gradnorm}. We provide a study case on this in Section \ref{sec:results}.

\begin{figure*}[!htp]
     \centering
    \begin{subfigure}[b]{0.32\textwidth}
        \includegraphics[width=\textwidth]{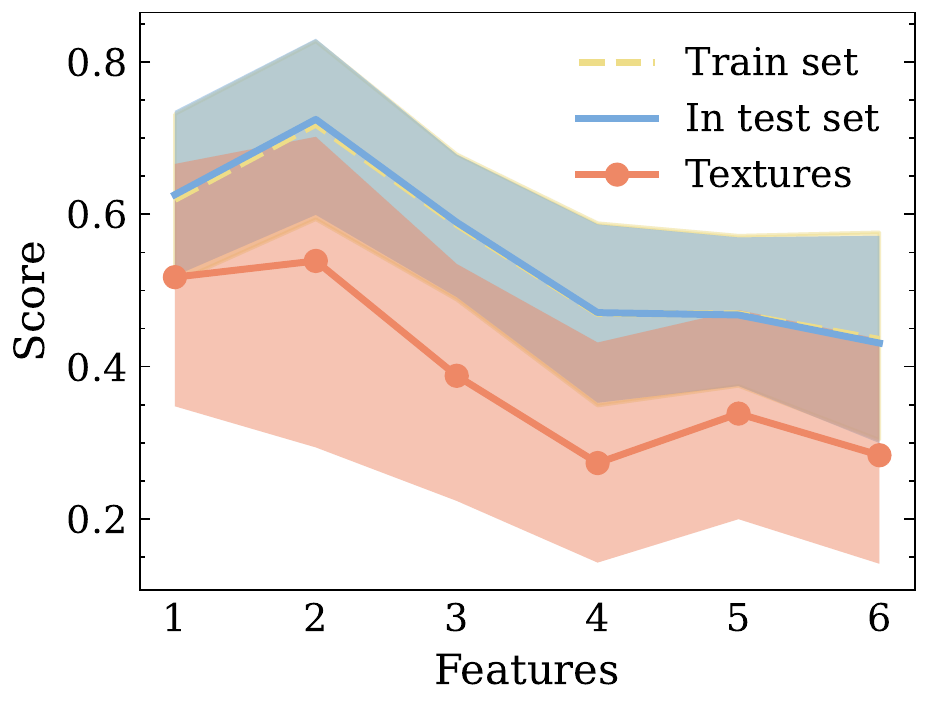}
        \caption{Functional trajectories.}
        \label{fig:vit_time_series_places}
    \end{subfigure}
    \hfill
    \begin{subfigure}[b]{0.32\textwidth}
        \centering
        \includegraphics[width=\textwidth]{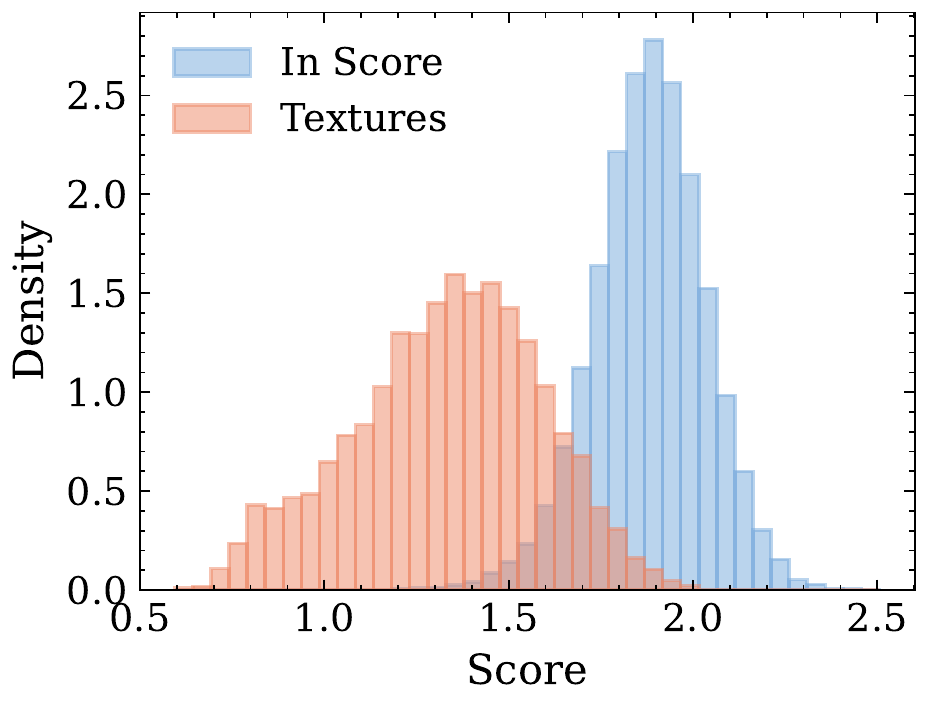}
        \caption{OOD score histogram.}
        \label{fig:vit_histogram_places}
    \end{subfigure}
    \hfill
    \begin{subfigure}[b]{0.32\textwidth}
        \centering
        \includegraphics[width=\textwidth]{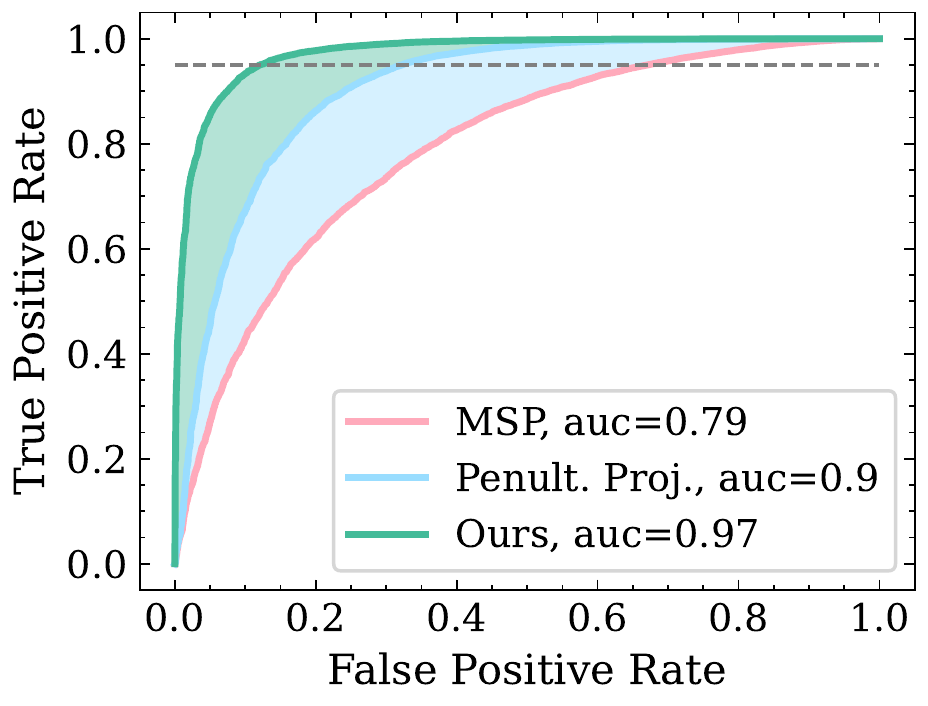}
        \caption{ROC curve.}
        \label{fig:vit_roc_places}
    \end{subfigure}
    \caption{Functional representation with 5 and 95\% quantiles (\ref{fig:vit_time_series_places}), histogram (\ref{fig:vit_histogram_places}), and ROC curve (\ref{fig:vit_roc_places}) for our OOD score on a DenseNet-121 model with Textures as OOD dataset.}
\end{figure*}

\textbf{Models.}
We ran experiments with five models. A \textit{DenseNet-121} \citep{densenet} pre-trained on ILSVRC-2012 with 8M parameters and test set top-1 accuracy of 74.43\%. A \textit{ResNet-50} model with top-1 test set accuracy of 75.85\% and 25M parameters. A \textit{BiT-S-101} \citep{Kolesnikov2020BigT} model based on a ResNetv2-101 architecture with top-1 test set accuracy of 77.41\% and 44M parameters. 
And a MobileNetV3 large, with accuracy of 74.6\% and around 5M parameters.
We reduced the intermediate representations with an \textit{max pooling} operation when needed obtaining a final vector with a dimension equal to the number of channels of each output.
We also ran experiments with a Vision Transformer (\textit{ViT-B-16}; \citealp{Dosovitskiy2021AnII}), which is trained on the ILSVRC2012 dataset with 82.64\% top-1 test accuracy and 70M parameters. We take the output's class tokens for each layer.
We download all the checkpoints weights from PyTorch \citep{pytorch} hub. All models are trained from scratch on ImageNet-1K. For further details, please refer to Appendix~\ref{ap:latent}. For all models, we compute the probability-weighted projection of the building blocks, as well as the projection of the logits of the network to form the functional representation. So, there is no need for a special layer selection.

\textbf{Evaluation Metrics.}\label{sec:eval_metrics}
We evaluate the methods in terms of {\AUROC} and {\TNR}. 
The Area Under The Receiving Operation Curve ({\AUROC}) is the area under the curve representing the true negative rate against the false positive rate when considering a varying threshold. It measures how well can the OOD score distinguish between out- and in-distribution data in a threshold-independent manner. 
The True Negative Rate at 95\% True Positive Rate (TNR at 95\% TPR or {\TNR} for short) is a threshold-dependent metric that provides the detector's performance in a reasonable choice of threshold. It measures the accuracy in detecting OOD samples when the accuracy of detecting in-distribution samples is fixed at 95\%. For both measures, higher is better. 

\textbf{Baseline methods.}
We followed the hyperparameter validation procedure suggested at the original papers for the baseline methods.
For ODIN \citep{odin}, we set the temperature to 1000 and the noise magnitude to zero. We take a temperature equal to one for Energy \citep{liu2020energybased}. We set the temperature to one for GradNorm  \citep{Huang2021OnTI_gradnorm}. For Mahalanobis \citep{mahalanobis}, we take only the scores of the outputs of the penultimate layer of the network. The MSP \citep{baseline} does not have any hyperparameters. For ReAct \citep{Sun2021ReActOD}, we compute the activation clipping threshold with a percentile equal to 90. For KNN \citep{Sun2022OutofdistributionDW} we set as hyperparameters $\alpha=1\%$ and $k=10$. 
%We didn't compare to further post-hoc methods as they would often depend from a specific architecture choice. 
\begin{table*}[!htp]\centering
\small
\caption{Comparison against post-hoc state-of-the-art methods for OOD detection on the ImageNet benchmark. M+Proj. stands for using the Mahalanobis distance-based layer score with our proposed unsupervised score aggregation algorithm based on trajectory similarity. Values are in percentage.}\label{tab:main}
\begin{tabular}{llcccccccc|cc}\toprule
& &\multicolumn{2}{c}{iNaturalist} &\multicolumn{2}{c}{SUN} &\multicolumn{2}{c}{Places} &\multicolumn{2}{c}{Textures} &\multicolumn{2}{c}{Average} \\\cmidrule{3-12}
& &\TNR &\ROC &\TNR &\ROC &\TNR &\ROC &\TNR &\ROC &\TNR &\ROC \\\midrule
\multirow{13}{*}{\rotatebox[origin=c]{90}{ResNet-50}} &
MSP  & 47.2  & 88.4  & 30.9  & 81.6  & 27.9  & 80.5  & 33.8  & 80.4  & 35.0  & 82.8 \\
&ODIN  & 58.9  & 91.3  & 35.4  & 84.7  & 31.6  & 82.0  & 49.5  & 84.9  & 43.9  & 85.9 \\
&Energy  & 46.3  & 90.6  & 41.2  & 86.6  & 34.0  & 84.0  & 47.6  & 86.7  & 42.3  & 87.0 \\ 
& MaxLogits  & 49.2  & 91.1  & 39.6  & 86.4  & 34.0  & 84.0  & 45.1  & 86.4  & 42.0  & 87.0 \\ 
& KLMatching  & 52.8  & 89.7  & 25.7  & 80.4  & 23.7  & 78.9  & 34.2  & 82.5  & 34.1  & 82.9 \\ 
& IGEOOD  & 42.2  & 90.1  & 34.7  & 85.0  & 29.8  & 82.8  & 43.5  & 85.7  & 37.6  & 85.9 \\ 
&  Mahalanobis  & 5.7  & 63.0  & 2.5  & 50.8  & 2.4  & 50.4  & 55.7  & 89.8  & 16.6  & 63.5 \\ 
&  GradNorm  & 73.2  & 93.9  & 62.6  & 90.1  & 51.1  & 86.1  & 67.2  & 90.6  & 63.5  & 90.2 \\ 
&   DICE  & 72.3  & 94.3  & 62.6  & 90.7  & 51.0  & 87.4  & 67.6  & 90.6  & 63.4  & 90.7 \\ 
&  ViM  & 28.3  & 87.4  & 17.9  & 81.0  & 16.7  & 78.3  & 85.2  & 96.8  & 37.0  & 85.9 \\ 
&  ReAct  & 82.2  & 96.7  & 74.9  & 94.3  & 65.4  & 91.9  & 48.7  & 88.8  & 67.8  & 93.0 \\ 
&   KNN  & 69.8  & 94.9  & 51.0  & 88.6  & 40.9  & 84.7  & 84.5  & 95.4  & 61.5  & 90.9 \\ 
\cmidrule{2-12}
    &   Proj. (Ours)  & 75.7  & 95.8  & 76.0  & 94.5  & 63.0  & 91.2  & 83.2  & 96.3  & \textbf{74.5}  & \textbf{94.5} \\
\midrule
\multirow{14}{*}{\rotatebox[origin=c]{90}{ViT-B-16}} 
&MSP  & 48.5  & 88.2  & 33.5  & 80.9  & 31.3  & 80.4  & 39.8  & 83.0  & 38.3  & 83.1 \\ 
&ODIN  & 49.9  & 86.0  & 31.5  & 75.2  & 33.7  & 76.5  & 42.6  & 81.2  & 39.4  & 79.7 \\ 
&Energy  & 35.9  & 79.2  & 27.2  & 70.2  & 25.7  & 68.4  & 41.5  & 79.3  & 32.6  & 74.3 \\ 
&MaxLogits  & 73.3  & 93.2  & 50.4  & 84.8  & 42.9  & 81.2  & 49.5  & 83.7  & 54.0  & 85.7 \\ 
& KLMatching  & 64.3  & 93.2  & 36.4  & 85.1  & 32.8  & 83.4  & 40.8  & 84.5  & 43.6  & 86.6 \\ 
& IGEOOD  & 79.0  & 94.6  & 55.0  & 85.9  & 46.3  & 81.8  & 55.0  & 85.0  & \textbf{58.8}  & 86.8 \\ 
& Mahalanobis  & 81.2  & 96.0  & 40.7  & 85.3  & 40.0  & 84.2  & 46.3  & 87.5  & 52.1  & 88.2 \\ 
& GradNorm  & 53.5  & 91.2  & 41.5  & 85.3  & 38.3  & 83.4  & 48.1  & 86.5  & 45.3  & 86.6 \\ 
&  ReAct  & 33.4  & 85.6  & 26.9  & 78.8  & 25.6  & 77.3  & 42.2  & 84.5  & 32.0  & 81.5 \\ 
&  DICE  & 2.8  & 46.1  & 3.2  & 49.3  & 4.5  & 49.7  & 19.0  & 64.3  & 7.4  & 52.3 \\ 
&  ViM  & 87.4  & 97.1  & 37.1  & 85.4  & 34.9  & 81.9  & 37.7  & 85.9  & 49.3  & 87.6 \\ 
& KNN  & 41.0  & 88.9  & 17.8  & 79.4  & 18.2  & 77.7  & 44.0  & 87.8  & 30.3  & 83.4 \\ 
\cmidrule{2-12}
& Proj. (Ours)  & 58.6  & 93.3  & 32.2  & 82.1  & 31.5  & 80.7  & 56.7  & 91.1  & 44.7  & 86.8 \\ 
&  M+Proj. (Ours)  & 77.9  & 95.5  & 34.8  & 83.2  & 32.7  & 81.4  & 79.1  & 94.9  & 56.1 & \textbf{88.8} \\ 
\bottomrule
\end{tabular}
\end{table*}
\section{Results and Discussion}\label{sec:results}
\textbf{Main results.} We report our main results in Table~\ref{tab:main}, which includes the performance for two out of five models (see~\cref{ap:additional_results} for the remaining three), four OOD datasets, and twelve detection methods.
% Our results are across the board consistent and on average superior to previous methods. 
On ResNet-50, we achieve a gain of 6.7\% in {\TNR} and 1.5\% in {\AUROC} compared to ReAct.
For the ViT-B-16, the gap between methods is small and our method exhibits a comparable {\TNR} and {\AUROC} to previous state-of-the-art.
For BiT-S-101, we outperform GradNorm by 18.9\% {\TNR} and 5.4\% {\AUROC}.
For DenseNet-121 (see~\cref{ap:additional_results}), we improved on ReAct by 16\% and 3.9\% in {\TNR} and {\AUROC}, respectively.
Finally, on MobileNet-V3 Large, we registered gains of around 20\% {\TNR} and 9.2\% {\AUROC}.
We observed that activation clipping benefits our method on convolution-based networks but hurts its performance on transformer architectures, aligned with the results from \cite{Sun2021ReActOD}. 

\begin{minipage}[c]{0.44\textwidth}
\centering
\small
\captionof{table}{Gain in terms of AUROC w.r.t two strong baselines (KNN and ViM) across five different models on the ImageNet benchmark.}\label{tab:diff_sota}
\resizebox{\columnwidth}{!}{
\begin{tabular}{lcc}
\toprule
Model  & Diff. KNN  & Diff. ViM \\ \hline
DenseNet-121  & +4.4\%  & +9.1\% \\ 
ResNet-50  & +0.9\%  & +6.0\% \\ 
ViT-B/16 (224)  & +5.4\%  & +1.2\% \\ 
ViT-B/16 (384)  & +0.8\%  & -0.9\% \\ 
MobileNetV3-Large  & +14.1\%  & +17\% \\ \bottomrule
\end{tabular}
}
\end{minipage}
\begin{minipage}[c]{0.54\textwidth}
\centering
\small
\captionof{table}{CIFAR-10 benchmark results in terms of {\AUROC} based on a ResNet-18 model.}\label{tab:results_cifar10}
\resizebox{\columnwidth}{!}{
\begin{tabular}{lccccc|c}\toprule
&MSP &ODIN &Energy &KNN &ReAct &Ours \\\midrule
C-100 &88.0 &88.8 &89.1 &89.8 &89.7 &89.4 \\
SVHN &91.5 &91.9 &92.0 &94.9 &94.6 &99.0 \\
LSUN (c) &95.1 &98.5 &98.9 &97.0 &97.9 &99.8 \\
LSUN (r) &92.2 &94.9 &95.3 &95.8 &96.7 &99.8 \\
TIN &89.8 &91.1 &91.7 &92.8 &93.8 &98.0 \\
Places &90.1 &92.9 &93.2 &93.7 &94.7 &93.6 \\
Textures &88.5 &86.4 &87.2 &94.2 &93.4 &97.9 \\
\midrule
Average &90.7 &92.1 &92.5 &94.0 &94.4 &\textbf{96.8} \\
\bottomrule
\end{tabular}
}
\end{minipage}

\textbf{Results on CIFAR-10.}
We ran experiments with a ResNet-18 model trained on CIFAR-10 \cite{cifar10}. We extracted the trajectory from the outputs of layers 2 to 4 and logits. The results are displayed in~\cref{tab:results_cifar10}. Our method outperforms comparable state-of-the-art methods by 2.4\% on average {\AUROC}, demonstrating that it is consistent and suitable for OOD detection on small datasets too.

\textbf{Multivariate OOD Scores is Not Enough.} 
Even though well-known multivariate novelty (or anomaly) detection techniques, such as One-class SVM \citep{cortes1995support}, Isolation Forest \citep{isolation2008forest}, Extended Isolation Forest \citep{Hariri2021ExtendedIF}, Local Outlier Factor \citep{lof}, k-NN approaches \citep{knn}, and distance-based approaches \citep{mahalanobis1936generalized} are adapted to various scenarios, they showed to be inefficient for integrating layers' information. 
%Unfortunately, classic multivariate methods cannot capture this interlayer dependence effectively.
A hypothesis that explains this failure is the important sequential dependence pattern we noticed in the in-distribution layer-wise scores. Table~\ref{tab:multivar_comb} shows the performance of a few unsupervised aggregation methods based on a multivariate OOD detection paradigm. We tried typical methods: evaluating the Euclidean and Mahalanobis distance w.r.t the training set Gaussian representation, fitting an Isolation Forest, and fitting a One-class SVM on training trajectory vectors. We compared the results with the performance of taking \textit{only} the penultimate layer scores, and we observed that the standard multivariate aggregation fails to improve the scores.

\begin{table*}[!htp]
\centering
\caption{The first row shows a single-layer baseline where \textit{Penultimate layer} is our layer score on the penultimate layer outputs. The subsequent rows compare the performance of unsupervised multivariate aggregation methods. We ran experiments with DenseNet-121.}\label{tab:multivar_comb}
\small
\begin{tabular}{lcccccccc|cc}\toprule
&\multicolumn{2}{c}{iNaturalist} &\multicolumn{2}{c}{SUN} &\multicolumn{2}{c}{Places} &\multicolumn{2}{c}{Textures} &\multicolumn{2}{c}{Average} \\\cmidrule{2-11}
&\TNR &\ROC &\TNR &\ROC &\TNR &\ROC &\TNR &\ROC &\TNR &\ROC \\\midrule
% Grad Norm &73.3 &93.4 &59.1 &88.8 &48.0 &84.1 &56.7 &87.7 &59.3 &88.5 \\
Penultimate layer (Ours) &\textbf{78.9} &\textbf{95.2} &61.9 &90.0 &51.6 &86.0 &68.8 &90.7 &65.3 &90.5 \\
\cmidrule{1-11}
Euclidean distance &63.7 &88.8 &53.6 &84.8 &40.8 &78.4 &83.5 &95.6 &60.4 &86.9 \\
Mahalanobis distance &40.1 &82.8 &26.6 &74.5 &20.8 &69.4 &73.5 &93.5 &40.3 &80.1 \\
Isolation Forest &60.3 &87.6 &46.9 &82.6 &36.3 &76.7 &81.0 &95.3 &56.1 &85.6 \\
One Class SVM &64.2 &89.0 &54.0 &85.0 &41.2 &78.6 &83.8 &95.7 &60.8 &87.0 \\
\cmidrule{1-11}
Trajectory Proj. (Ours) &65.7 &92.8 &\textbf{68.0} &\textbf{92.1} &\textbf{52.4} &\textbf{87.3} &\textbf{88.3} &\textbf{97.5} &\textbf{68.6} &\textbf{92.4} \\
\bottomrule
\end{tabular}
\end{table*}

\textbf{Qualitative Evaluation of the Functional Dataset.}  The test in-distribution trajectories follow a well-defined trend similar to the training distribution (see~\cref{fig:vit_time_series_places}). While the OOD trajectories manifest mainly as shape and magnitude anomalies (w.r.t. the taxonomy of \citealp{HubertRS15,staerman2022functional}). These characteristics reflect on the histogram of our detection score (see~\cref{fig:vit_histogram_places}). The in-distribution histogram is generally symmetric, while the histogram for the OOD data is typically skewed with a smaller average. Please refer to~\cref{ap:further_images} for additional figures.

\begin{figure}[!htp]
     \centering
    \begin{subfigure}[b]{0.48\textwidth}\includegraphics[width=\textwidth]{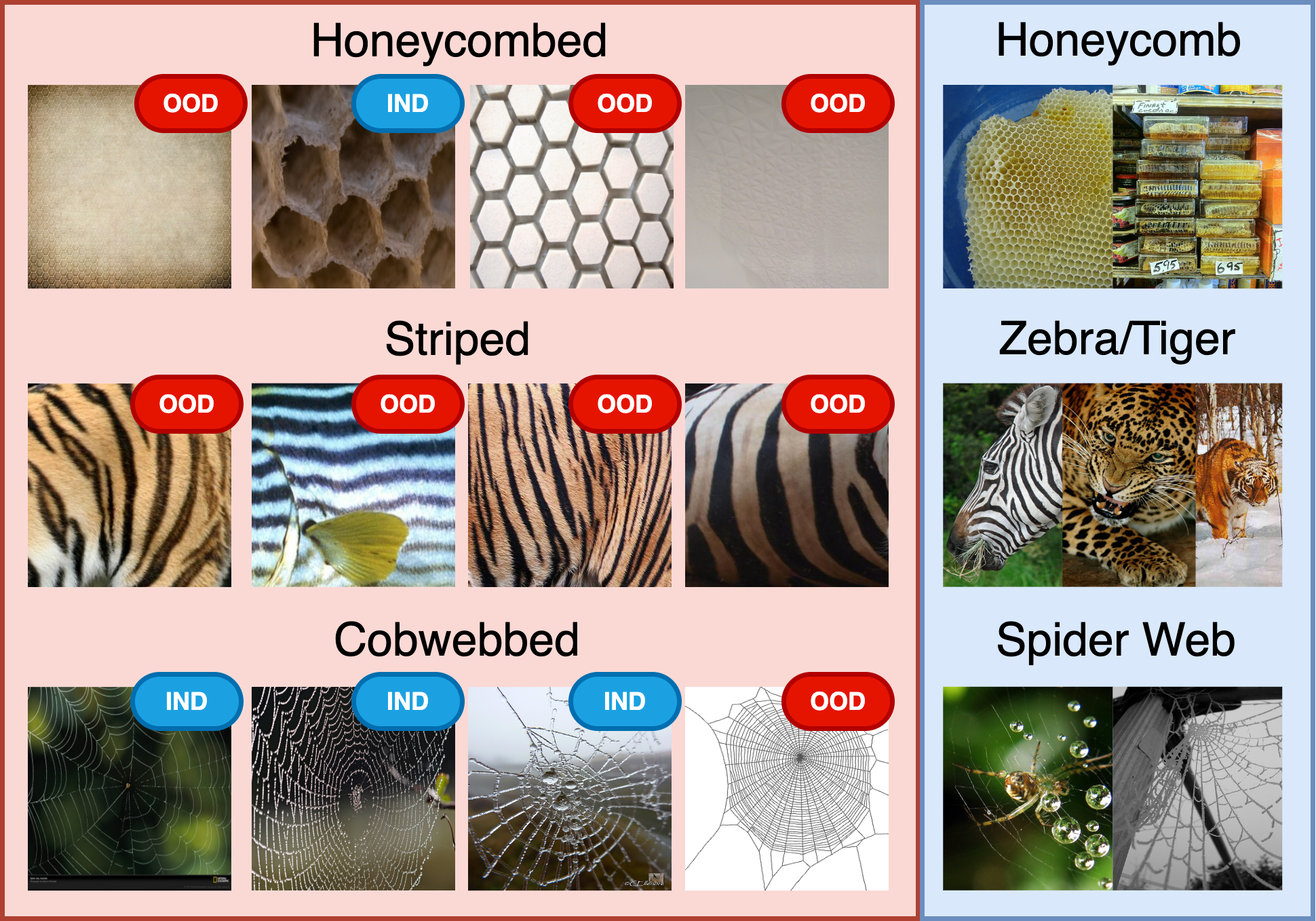}
        \caption{A few examples from Textures dataset sharing semantic overlap with ImageNet classes.}
        \label{fig:cherry_1}
    \end{subfigure}
    % \hspace{1cm}
    \hfill
    \begin{subfigure}[b]{0.48\textwidth}
        \centering
        \includegraphics[width=\textwidth]{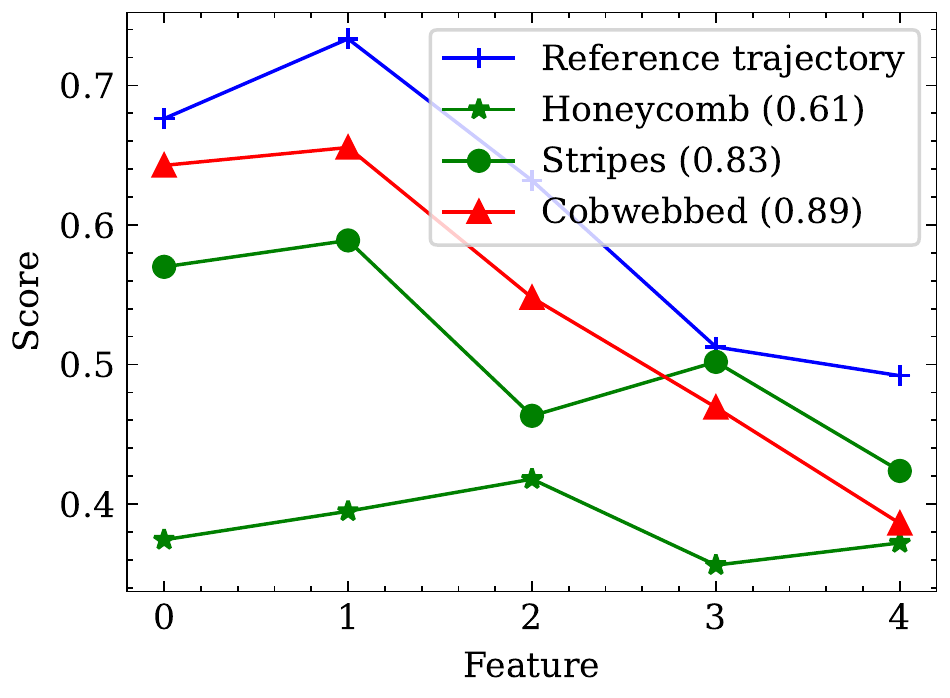}
        \caption{Trajectories of the leftmost examples of Fig. \ref{fig:cherry_1} and their OOD scores in parenthesis.}
        \label{fig:cherry_2}
    \end{subfigure}
    \caption{Detection of individual samples on classes with semantic overlap between ImageNet and Textures. The badge on each image on Fig.~\ref{fig:cherry_1} shows the detection label given by our method.
    % We set as threshold the score value with 95\% TPR.
    }
\end{figure}

\textbf{Study Case.} There are a few overlaps in terms of the semantics of class names in the Textures and ImageNet datasets. In particular, ``honeycombed" in Textures versus ``honeycomb" in ImageNet, ``stripes" vs. ``zebra", "tiger", and "tiger cat", and ``cobwebbed" vs. ``spider web". We showed in~\cref{tab:main} that our method significantly decreases the number of false negatives in this benchmark. In order to better understand how our method can discriminate where baselines often fail, we designed a simple study case. Take the Honeycombed vs. Honeycomb, for instance (the first row of Fig.~\ref{fig:cherry_1}). The honeycomb from ImageNet references natural honeycombs, usually found in nature, while honeycombed in Textures has a broader definition attached to artificial patterns. In this class, the Energy baseline makes 108 mistakes, while we only make 20 mistakes. We noticed that some of our mistakes are aligned with real examples of honeycombs (e.g., the second example from the first row), whilst we confidently classify other patterns correctly as OOD. For the striped case (middle row), our method flags only 16 examples as being in-distribution, but we noticed an average higher score for the trajectories in Fig. \ref{fig:cherry_2} (Stripes). Note that, for the animal classes, the context and head are essential features for classifying them. For the Spider webs class, most examples from Textures are visually closer to ImageNet. Overall, the study shows that our scores are aligned with the semantic proximity between testing samples and the training set. 

\textbf{Limitations.} 
We believe this work is only the first step towards efficient post-score aggregation as we have tackled an open and challenging problem of combining multi-layer information. We suspect there is room for improvement since our metric lives in the inner product space, which is a specific case for more general structures found in Hilbert spaces.
Also, we rely on the class-conditional mean vectors, which might not be sufficiently informative for statistically modeling the  embedding features depending on the distribution. From a statistical point of view, the average would be informative if the data is compact. To address this point, we plotted the median and the mean for several coordinates of the feature map and measured their difference in~\cref{fig:mean_median_hist}. We observed that they practically superpose in most dimensions, which indicates that the data is compact and central, thus, the prototypes are informative. Additionally, we showed in~\cref{fig:wrap_mean_depth} that the halfspace depth (\cite{Tukey1975}; more details in~\cref{ap:centrality_features}) of the mean vector of a given class is superior to the maximum depth of a training sample of the same class, suggesting that the average is deep in the feature's data distribution. 
% in the embedding features.
% , especially as more accurate classifiers are developed.

% Another concern that may arise is that, from a practical point of view, current third-party ML services often restrict the practitioner from accessing the intermediate outputs of the model in production. Nonetheless, the service provider has access to this information and could leverage it to deliver OOD detection as a service, for instance. 
%We hope our research will be useful for inspiring new techniques and works towards better OOD detection 

\textbf{Broader Impacts.}
Our goal is to improve the trustworthiness of modern machine learning models by creating a tool that can detect OOD samples. Hence, we aim to minimize the uncertainties and dangers that arise when deploying these models. Although we are confident that our efforts will not lead to negative consequences, caution is advised when using detection methods in critical domains.
% it is crucial to be cautious when using detection methods in critical domains.

\section{Conclusion}\label{sec:conclusion}

In this work, we introduced an original approach to OOD detection based on a functional view of the pre-trained multi-layer neural network that leverages the sample’s trajectories through the layers.
Our method detects samples whose trajectories differ from the typical behavior characterized by the training set. The key ingredient relies on the statistical dependencies of the scores extracted at each layer, using a purely self-supervised algorithm. We validate empirically, through an extensive benchmark against several state-of-the-art methods, that our Projection method is consistent and achieves great results across multiple models and datasets, even though requiring no special hyperparameter tuning. We hope this work will encourage future research to explore sample trajectories to enhance the safety of AI systems.
%Beyond the novelty and practical advantages of the algorithm, our results establish the value of using a functional approach as an unsupervised technique to combine multiple scores, which offers an exciting alternative to usual single-layer detection methods.

%by encouraging further studies to look into sample's trajectories through a modern neural network.

\ack{This work has been supported by the project PSPC AIDA: 2019-PSPC-09 funded by BPI-France and was granted access to the HPC/AI resources of IDRIS under the allocation 2022 - AD011012803R1 made by GENCI.}

\bibliography{biblio}
\bibliographystyle{plain}

%%%%%%%%%%%%%%%%%%%%%%%%%%%%%%%%%%%%%%%%%%%%%%%%%%%%%%%%%%%%%%%%%%%%%%%%%%%%%%%
%%%%%%%%%%%%%%%%%%%%%%%%%%%%%%%%%%%%%%%%%%%%%%%%%%%%%%%%%%%%%%%%%%%%%%%%%%%%%%%
% APPENDIX
%%%%%%%%%%%%%%%%%%%%%%%%%%%%%%%%%%%%%%%%%%%%%%%%%%%%%%%%%%%%%%%%%%%%%%%%%%%%%%%
%%%%%%%%%%%%%%%%%%%%%%%%%%%%%%%%%%%%%%%%%%%%%%%%%%%%%%%%%%%%%%%%%%%%%%%%%%%%%%%
\newpage
\appendix
\onecolumn

\section{Appendix}
\subsection{Algorithm and Computational Details}\label{sec:alg_appendix}

This section introduces further details on the computation algorithm and resources. Algorithms~\ref{alg:neural_fun_rep} and \ref{alg:test_ood_score} describe how to extract the neural functional representations from the samples and compute the OOD score from test samples, respectively. Note that we emphasize the ``functional representations", because the global behavior of the trajectory matters. In very basic terms, the method can look into the past and future of the series, contrary to a ``sequential point of view" which is restricted to the past of the series only, not globally on the trajectory.

% \begin{algorithm}[tb]
%   \caption{Bubble Sort}
%   \label{alg:example}
% \begin{algorithmic}
%   \STATE {\bfseries Input:} data $x_i$, size $m$
%   \REPEAT
%   \STATE Initialize $noChange = true$.
%   \FOR{$i=1$ {\bfseries to} $m-1$}
%   \IF{$x_i > x_{i+1}$}
%   \STATE Swap $x_i$ and $x_{i+1}$
%   \STATE $noChange = false$
%   \ENDIF
%   \ENDFOR
%   \UNTIL{$noChange$ is $true$}
% \end{algorithmic}
% \end{algorithm}

\begin{algorithm}[ht]
\caption{Neural functional representation extraction algorithm computed offline.}
\label{alg:neural_fun_rep}
\begin{algorithmic}
\STATE {\bfseries Input:} Training dataset $\mathcal{S}_N=\{\left(\boldsymbol{x}_i, y_i \right)\}_{i=1}^{N}$ and a pre-trained DNN $f=f_{L+1} \circ \dots \circ f_1$.
\STATE  \texttt{// Training dataset feature extraction}
\FOR{$\ell \in \{1,\ldots, L+1\}$}
\STATE $\boldsymbol{z}_{\ell,i} \gets (f_{\ell} \circ \dots \circ f_1)(\boldsymbol{x}_i)$ 
\FOR{$y \in \mathcal{Y}$}
\STATE  $\boldsymbol{\mu}_{\ell,y} \gets \frac{1}{N_y}\sum_{i=1}^{N_y} \boldsymbol{z}_{\ell,i} \quad $ \texttt{// Class conditional features prototypes}
\ENDFOR
\ENDFOR
\FOR{$i \in \{1, \ldots, N\}$}
\STATE \texttt{// Functional trajectory extractor $\phi(\cdot)$}
\STATE $\boldsymbol{u}_{i} \gets \left[ {\rm{d}_1}(\boldsymbol{z}_{1,i};\{\boldsymbol{\mu}_{1,y} : \forall y \in \mathcal{Y} \}),\dots,{\rm{d}_{L+1}}(\boldsymbol{z}_{L+1,i};\{\boldsymbol{\mu}_{L+1,y} : \forall y \in \mathcal{Y} \}) \right]$ 
\ENDFOR
\STATE $\mathcal{U} \gets \{\boldsymbol{u}\}_{i=1}^{N}$
\STATE $\max\left(\mathcal{U}\right) \gets [\max_i {u_{i,1}}, \dots, \max_i u_{i,L+1}]^\top $
\STATE $\bar{\boldsymbol{u}} = \frac{1}{N} \sum_{i=1}^N \frac{\boldsymbol{u}_i}{\max\left(\mathcal{U}\right)}$
\STATE {\bfseries Return:} $\bar{\boldsymbol{u}}$, $\max\left(\mathcal{U}\right)$, $\phi(\cdot;\{\boldsymbol{\mu}_{\ell,y} :\  y \in \mathcal{Y} \text{ and } \ell \in \{1,\dots,L+1\}\})$
\end{algorithmic}
\end{algorithm}

\begin{algorithm}[ht]
\caption{Out-of-distribution score computation online.}
\label{alg:test_ood_score}
\begin{algorithmic}
\STATE {\bfseries Input:} Test sample $\boldsymbol{x}_0$, the DNN $f=f_{L+1} \circ \dots \circ f_1$, reference trajectory $\bar{\boldsymbol{u}}$, scaling vector $\max\left(\mathcal{U}\right)$, and neural functional trajectory extraction function $\phi(\cdot;\{\boldsymbol{\mu}_{\ell,y} :\  y \in \mathcal{Y} \text{ and } \ell \in \{1,\dots,L+1\}\})$.
\FOR{$\ell \in \{1,\ldots, L+1\}$}
\STATE $\boldsymbol{z}_{\ell,0} \gets (f_{\ell} \circ \dots \circ f_1)(\boldsymbol{x}_0)$ 
\ENDFOR
\STATE $\boldsymbol{u}_0 \gets \phi(\boldsymbol{z}_0;\{\boldsymbol{\mu}_{\ell,y} : y \in \mathcal{Y}\text{ and } \ell \in \{1,\dots,L+1\}\})$
\STATE $\tilde{\boldsymbol{u}}_0 \gets \frac{\boldsymbol{u}_0}{\max(\mathcal{U})}$
\STATE $s(\boldsymbol{x}_0) \gets \frac{1}{\lVert\bar{\boldsymbol{u}}\rVert^2}\sum_{i=1}^{L+1} \tilde{\boldsymbol{u}}_{0,i} \bar{\boldsymbol{u}}_i$
\STATE {\bfseries Return:} $s(\boldsymbol{x}_0)$
\end{algorithmic}
\end{algorithm}

\subsubsection{Computing Resources}\label{sec:computing_appendix}

We run our experiments on an internal cluster. Since we use pre-trained popular models, it was not necessary to retrain the deep models. Thus, our results should be reproducible with a single GPU. Since we are dealing with ImageNet datasets (approximately 150GB), a large storage capacity is expected. We save the features in memory to speed up downstream analysis of the algorithm, which may occupy around 200GB of storage.

\subsubsection{Inference time}\label{sec:time_analysis}

In this section, we conduct a time analysis of our algorithm. It is worth noting that most of the calculation burden is done offline. At inference, only a forward pass and feature-wise scores are computed. We conducted a practical experiment where we performed live inference and OOD computation with three models for the MSP, Energy, Mahalanobis, and Trajectory Projection (ours) methods. The results normalized by the inference time are available in Table \ref{tab:time_analysis} below. We reckon that there may exist better computationally efficient implementations of these algorithms. So this remains a naive benchmark of their computational overhead.
\begin{table}[!htp]\centering
\caption{Batch runtime for OOD detection methods normalized by the time of one forward pass.}\label{tab:time_analysis}
\small
\begin{tabular}{lccccc}\toprule
&Forward pass &MSP &Energy &Mahalanobis &Ours \\\midrule
ResNet-50 &1.00 &1.00 &1.00 &1.18 &1.15 \\
BiT-S-101 &1.00 &1.00 &1.00 &1.21 &1.19 \\
DenseNet-121 &1.00 &1.00 &1.01 &1.54 &1.61 \\
ViT-B-16 &1.00 &1.01 &1.05 &2.12 &2.15 \\
\bottomrule
\end{tabular}
\end{table}

\subsection{Additional Results}\label{ap:additional_results}

In this section, we provide additional results in terms of {\TNR} and {\AUROC} for the BiT-S-101, DenseNet-121, and MobileNetV3-Large models in \cref{tab:main2} for the baseline methods and ours.

\begin{table}[!htp]\centering
% \small
\caption{Comparison against post-hoc state-of-the-art methods for OOD detection on the ImageNet benchmark. Values are in percentage.}\label{tab:main2}
\begin{tabular}{lccccccccc|cc}\toprule
& &\multicolumn{2}{c}{iNaturalist} &\multicolumn{2}{c}{SUN} &\multicolumn{2}{c}{Places} &\multicolumn{2}{c}{Textures} &\multicolumn{2}{c}{Average} \\\cmidrule{3-12}
& &\TNR &\ROC &\TNR &\ROC &\TNR &\ROC &\TNR &\ROC &\TNR &\ROC \\\midrule
\multirow{12}{*}{\rotatebox[origin=c]{90}{BiT-S-101}} &MSP &35.9 &87.9 &28.8 &81.9 &21.3 &79.3 &21.9 &77.5 &27.0 &81.7 \\
&ODIN &29.3 &86.7 &36.7 &86.8 &26.0 &82.7 &24.1 &79.3 &29.0 &83.9 \\
&Energy &25.0 &84.5 &39.8 &87.3 &27.2 &82.7 &24.2 &78.8 &29.0 &83.3 \\
& MaxLogits & 80.8 & 95.9 & 35.8 & 80.2 & 32.3 & 76.8 & 33.9 & 80.6 & 45.7 & 83.4 \\ 
& KLMatching & 68.5 & 92.9 & 28.6 & 81.5 & 27.4 & 80.2 & 38.0 & 84.2 & 40.6 & 84.7 \\ 
& IGEOOD & \textbf{83.8} & \textbf{96.4} & 37.6 & 82.6 & 33.4 & 79.3 & 36.8 & 82.9 & 47.9 & 85.3 \\ 
&Mahalanobis &16.5 &78.3 &13.2 &74.5 &10.5 &69.6 &\textbf{86.6} &\textbf{97.3} &31.7 &79.9 \\
&GradNorm &41.3 &86.0 &55.2 &88.2 &39.0 &\textbf{83.3} &41.2 &81.0 &44.2 &84.6 \\
&ReAct &46.2 &88.9 &10.7 &65.9 &7.0 &62.0 &8.5 &65.8 &18.1 &70.7 \\
& DICE & 16.3 & 86.4 & 8.2 & 69.1 & 4.6 & 61.9 & 5.2 & 62.9 & 8.6 & 70.1 \\ 
& ViM & 16.3 & 86.4 & 8.2 & 69.1 & 4.6 & 61.9 & 5.2 & 62.9 & 8.6 & 70.1 \\ 
%&DICE &83.6 &96.3 &36.5 &79.5 &30.3 &75.0 &29.1 &78.9 &44.9 &82.4 \\
&KNN &39.7 &88.9 &22.1 &77.5 &20.6 &75.9 &54.1 &89.7 &34.1 &83.0 \\
\cmidrule{2-12}
&Proj. (Ours) & 67.0 &91.7 &\textbf{59.5} &\textbf{89.4} &\textbf{40.8} &82.3 &85.1 &96.7 &\textbf{63.1} &\textbf{90.0 }\\
\cmidrule{1-12}
\multirow{13}{*}{\rotatebox[origin=c]{90}{DenseNet-121}} 
&MSP &50.7 &89.1 &33.0 &81.5 &30.8 &81.1 &32.9 &79.2 &36.9 &82.7 \\
&ODIN &60.4 &92.8 &45.2 &87.0 &40.3 &85.1 &45.3 &85.0 &47.8 &87.5 \\
&Energy &60.3 &92.7 &48.0 &87.4 &42.2 &85.2 &47.9 &85.4 &49.6 &87.7 \\
& MaxLogits & 58.2 & 92.3 & 44.6 & 86.8 & 38.7 & 84.5 & 45.1 & 84.8 & 46.6 & 87.1 \\ 
& KLMatching & 50.0 & 89.6 & 22.7 & 79.8 & 23.6 & 78.8 & 36.1 & 82.4 & 33.1 & 82.7 \\ 
& IGEOOD & 46.7 & 90.3 & 37.4 & 84.8 & 32.2 & 82.6 & 41.6 & 83.9 & 39.5 & 85.4 \\ 
&Mahalanobis &3.5 &59.7 &4.8 &57.0 &4.6 &54.8 &54.4 &88.3 &16.8 &64.9 \\
&GradNorm &73.3 &93.4 &59.1 &88.8 &48.0 &84.1 &56.7 &87.7 &59.3 &88.5 \\
&ReAct &68.8 &93.9 &51.3 &89.6 &44.5 &86.6 &48.1 &87.6 &53.2 &89.4 \\
&DICE & 72.0 & 93.9 & 61.0 & 89.8 & 48.8 & 85.8 & 58.6 & 87.8 & 60.1 & 89.3 \\ 
& ViM & 27.4 & 85.8 & 14.3 & 77.6 & 11.7 & 73.8 & 80.9 & 96.1 & 33.6 & 83.3 \\ 
% &DICE &61.9 &92.9 &49.3 &87.6 &42.9 &85.2 &49.8 &85.9 &51.0 &87.9 \\
&KNN &57.1 &92.1 &33.2 &83.6 &26.8 &79.6 &81.5 &\textbf{96.5} &49.7 &88.0 \\
\cmidrule{2-12}
% &Proj (Ours) &65.7 &92.8 &\textbf{68.0} &\textbf{92.1} &\textbf{52.4} &87.3 &\textbf{88.3} &\textbf{97.5} &68.6 &92.4 \\
&Proj. (Ours) &\textbf{80.4} &\textbf{96.4} &\textbf{62.2} &\textbf{91.8} &\textbf{52.3} &\textbf{88.0} &\textbf{81.8} &\textbf{96.5} &\textbf{69.2} &\textbf{93.2}\\
\midrule
\multirow{13}{*}{\rotatebox[origin=c]{90}{MobileNetV3 Large}}
& MSP  & \textbf{39.4}  & \textbf{87.3}  & 25.0  & 79.9  & 23.8  & 79.0  & 29.7  & 80.8  & 29.5  & 81.8 \\ 
& ODIN  & 39.1  & 86.9  & 26.1  & 78.8  & 23.4  & 77.5  & 31.5  & 80.6  & 30.0  & 80.9 \\ 
& Energy  & 30.1  & 83.9  & 21.2  & 76.7  & 18.1  & 74.8  & 28.7  & 78.9  & 24.5  & 78.6 \\ 
& MaxLogits & 39.4 & 87.0 & 26.2 & 78.9 & 23.5 & 77.6 & 31.6 & 80.7 & 30.2 & 81.0 \\
& KLMatching & 46.6 & 89.4 & 19.8 & 78.7 & 20.8 & 77.7 & 31.9 & 83.2 & 29.7 & 82.2 \\
& IGEOOD & 35.4 & 86.8 & 24.8 & 79.2 & 21.8 & 77.6 & 32.6 & 81.5 & 28.7 & 81.3 \\
& Mahalanobis  & 7.3  & 71.3  & 2.2  & 52.7  & 3.3  & 54.4  & 45.7  & 84.7  & 14.6  & 65.8 \\ 
& GradNorm  & 11.6  & 76.2  & 7.9  & 68.3  & 9.3  & 68.7  & 14.6  & 76.7  & 10.9  & 72.5 \\ 
& ReAct  & 15.0  & 82.0  & 8.4  & 70.7  & 10.2  & 70.5  & 38.6  & 85.1  & 18.1  & 77.1 \\ 
& DICE & 23.2 & 83.5 & 12.4 & 76.8 & 14.0 & 75.8 & 13.3 & 78.0 & 15.7 & 78.5 \\
& ViM & 8.0 & 75.1 & 4.4 & 61.5 & 5.4 & 63.7 & 34.6 & 83.1 & 13.1 & 70.8 \\ 
& KNN  & 31.3  & 78.8  & 24.9  & 77.9  & 18.5  & 73.5  & 12.8  & 64.5  & 21.9  & 73.7 \\
\cmidrule{2-12}
& Proj. (Ours)  & 34.1  & 86.3  & \textbf{49.4}  & \textbf{87.1}  & \textbf{37.4}  & \textbf{82.4}  & \textbf{80.1}  & \textbf{95.4}  & \textbf{50.3}  & \textbf{87.8} \\
\bottomrule
\end{tabular}
\end{table}
\subsection{Latent Features}\label{ap:latent}
We list below the node names of the latent features extracted from each of the models trained on ImageNet-1K.
\subsubsection{ResNet-50}
\begin{small}
\texttt{[layer1, layer2, layer3, layer4, flatten, fc]}
\end{small}

\subsubsection{BiT-S-101}
% We used the outputs of the following features to calculate our scores for the BiT-S-101 based on a ResNet-101-v2 model:

\begin{small}
\texttt{[body.block1, body.block2, body.block3, body.block4, head.flatten]}
\end{small}

\subsubsection{DenseNet-121}
% We used the outputs of the following features to calculate our scores for the DenseNet-121 model:

\begin{small}
\texttt{[features.transition1.pool, features.transition2.pool, features.transition3.pool, features.norm5, flatten, classifier]}
\end{small}

\subsubsection{ViT-B-16}
% We used the outputs of the following features to calculate our scores for the ViT-B-16 model:

\begin{small}
\texttt{[encoder.layers.encoder\_layer\_0, encoder.layers.encoder\_layer\_1, encoder.layers.encoder\_layer\_2, encoder.layers.encoder\_layer\_3, encoder.layers.encoder\_layer\_4, encoder.layers.encoder\_layer\_5, encoder.layers.encoder\_layer\_6, encoder.layers.encoder\_layer\_7, encoder.layers.encoder\_layer\_8, encoder.layers.encoder\_layer\_9, encoder.layers.encoder\_layer\_10, encoder.layers.encoder\_layer\_11, encoder.ln, getitem\_5, heads.head]}
\end{small}

\subsection{MobileNetV3-Large}
\begin{small}
\texttt{[blocks.0, blocks.1, blocks.2, blocks.3, blocks.4, blocks.5, blocks.6, flatten, classifier]}
\end{small}

\subsubsection{Intra-block Convolutional Layers Ablation Study}

\begin{figure}[ht]
    \centering
    \begin{subfigure}[b]{0.24\textwidth}
        \includegraphics[width=\textwidth]{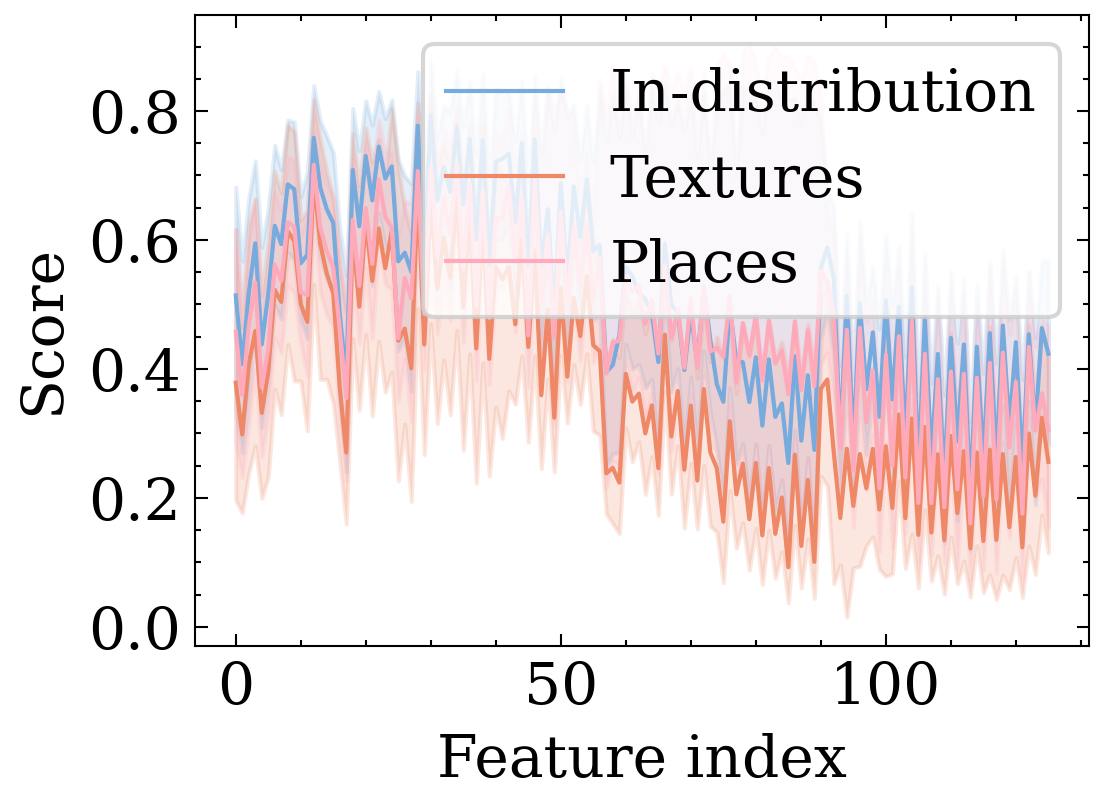}
        \caption{DenseNet-121 model.}
        \label{fig:conv_dn_1}
    \end{subfigure}
    \hfill
    \begin{subfigure}[b]{0.24\textwidth}
        \centering
        \includegraphics[width=\textwidth]{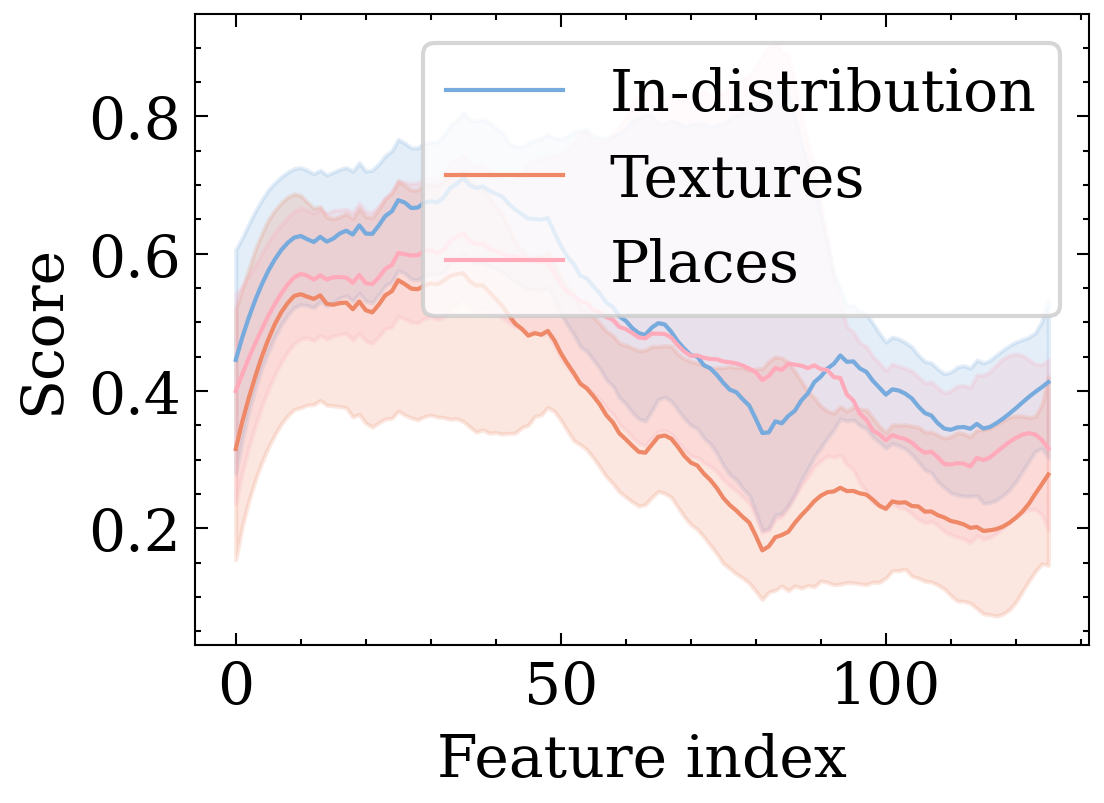}
        \caption{Smoothed trajectories.}
        \label{fig:conv_dn_2}
    \end{subfigure}
    \hfill
    \begin{subfigure}[b]{0.24\textwidth}
        \centering
\includegraphics[width=\textwidth]{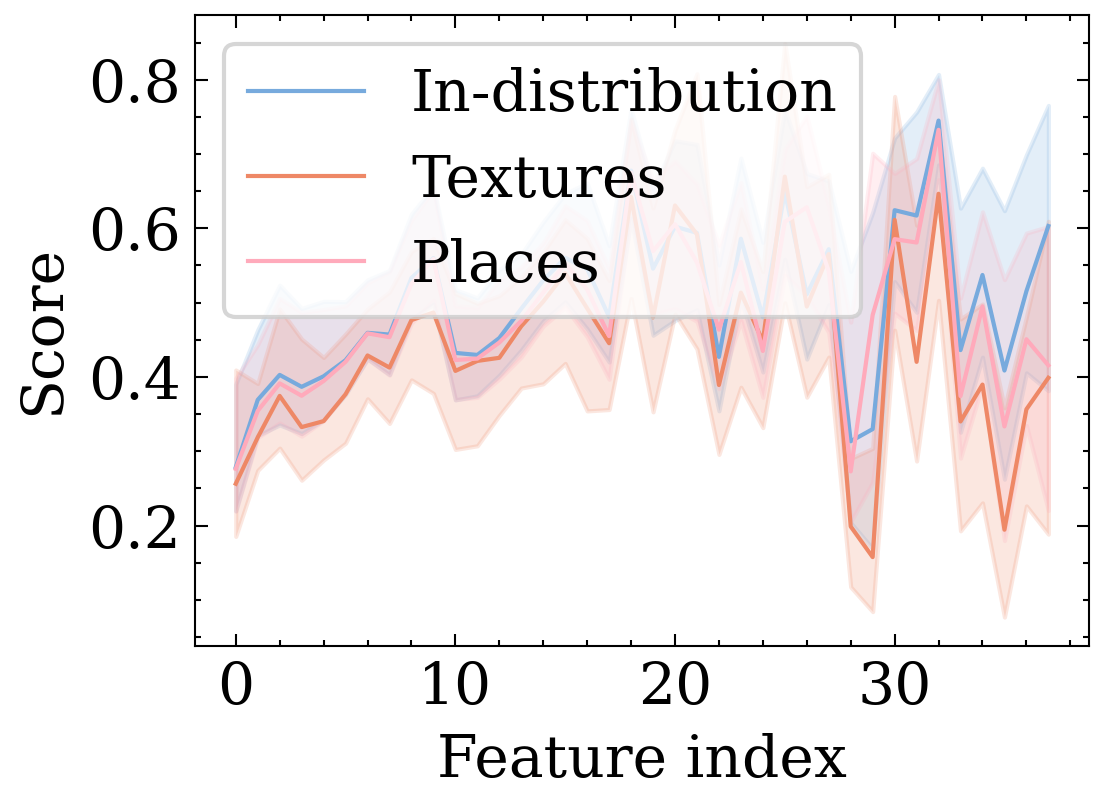}
        \caption{ResNet-50 model.}
        \label{fig:conv_bit_1}
    \end{subfigure}
    \hfill
    \begin{subfigure}[b]{0.24\textwidth}
        \includegraphics[width=\textwidth]{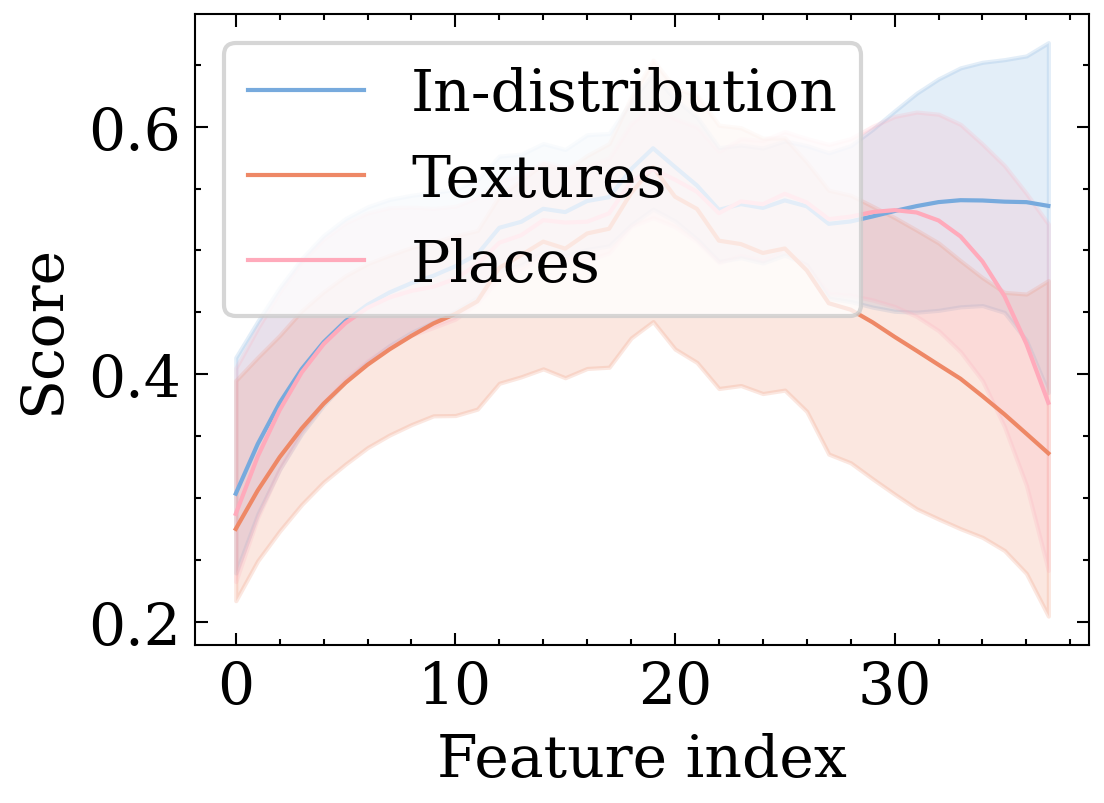}
        \caption{Smoothed trajectories.}
        \label{fig:conv_bit_2}
    \end{subfigure}
    \hfill
       \caption{Trajectories of scores from every intermediate convolutional layer.}
    \label{fig:all_conv_features}
\end{figure}

We showed through several benchmarks that taking the outputs of each convolutional block for the DenseNet-121 and ResNet-50 models is enough to obtain excellent results. We conduct further ablation studies to understand the impact of adding the intra-blocks convolutional layers in the trajectories. To do so, we extracted the outputs of every intermediate convolutional output of these networks, and we plotted their score trajectories in Figure~\ref{fig:all_conv_features}. The resulting trajectories are noisy and would be unfitting to represent the data properly. As a solution, we propose smoothing the curves with a polynomial filter to extract more manageable trajectories. Also, we observe regions of high variance for the in-distribution trajectory inside block 2 for the DenseNet-121 model (see Fig. \ref{fig:conv_dn_2}). A further preprocessing could be simply filtering out these features of high variance that would be unreliable for the downstream task of OOD detection.

\subsection{On the Centrality of the Class-Conditional Features Maps}\label{ap:centrality_features}

\begin{figure}
\centering
\includegraphics[width=5.5cm]{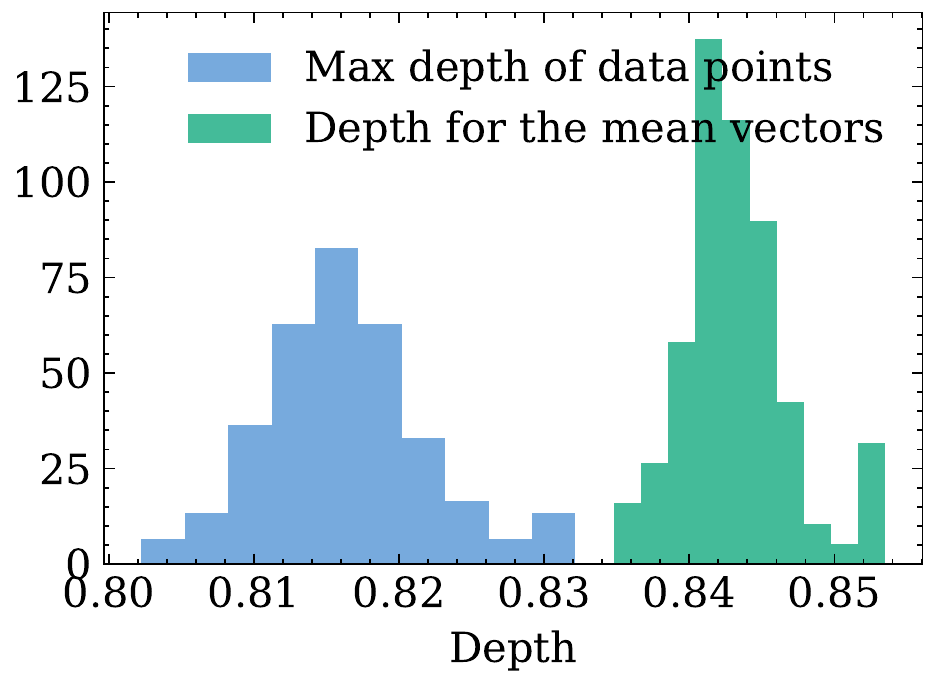}
\caption{Histogram showing that the halfspace depth of the average vectors for a given class is higher than the highest depth of an embedding feature vector of the same class, demonstrating multivariate centrality.}\label{fig:wrap_mean_depth}
\end{figure}
\begin{figure}[!ht]
    \centering
    \includegraphics[width=\textwidth]{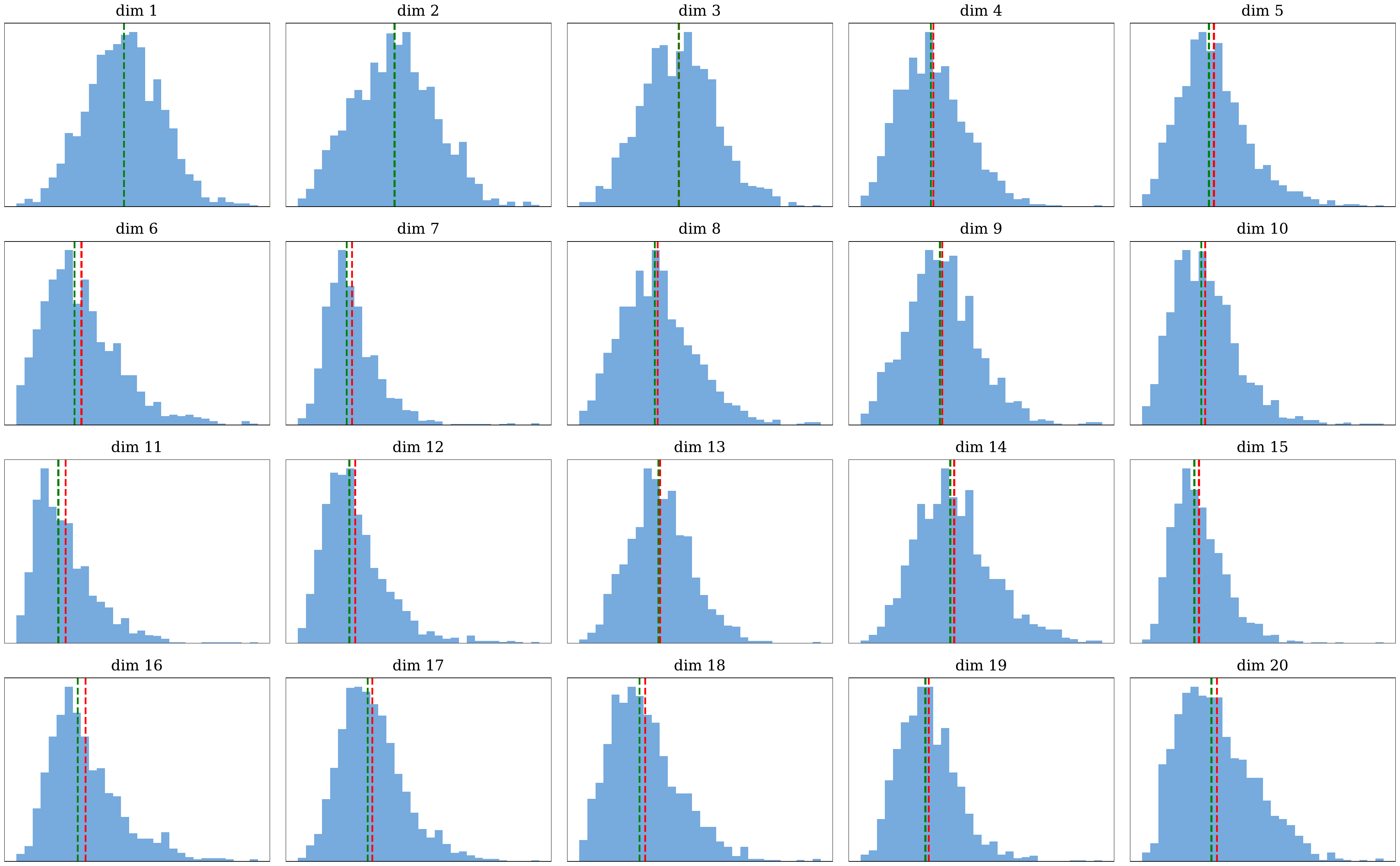}
    \caption{Histogram for the 20 first dimensions of the penultimate feature of a DenseNet-121 for class index 0 of ImageNet. The green line is the average, and the red line is the estimated median.}
    \label{fig:mean_median_hist}
\end{figure}

 
In this section, we study whether the mean vectors are sufficiently informative for statistically modeling the class-conditional embedding features. From a statistical point of view, the average would be informative if the data is compact. To address this point, we plotted the median and the mean for the coordinates of the feature map and measured their difference in Figure~\ref{fig:mean_median_hist}. We observed that they almost superpose in most dimensions or are separated by a minor difference, which indicates that the data is compact and central. In addition, we showed in Figure \ref{fig:wrap_mean_depth} that the halfspace depth \citep{Tukey1975} (see also, e.g. \cite{phdguigui}, Chapter 2 for a review of data depth) of the mean vector of a given is superior to the maximum depth of a training sample vector of the same class, suggesting the average is central or deep in the feature data distribution. From a practical point of view, the clear advantage of using only the mean as a reference are computational efficiency, simplicity, and interpretability. We believe that future work directions could be exploring a method that better models the density in the embedding features, especially as more accurate classifiers are developed.

\subsection{Additional Plots, Functional Dataset, Histograms and ROC Curves}\label{ap:further_images}

\begin{figure}[ht]
    \centering
    
    \begin{subfigure}[b]{0.329\textwidth}
        \includegraphics[width=\textwidth]{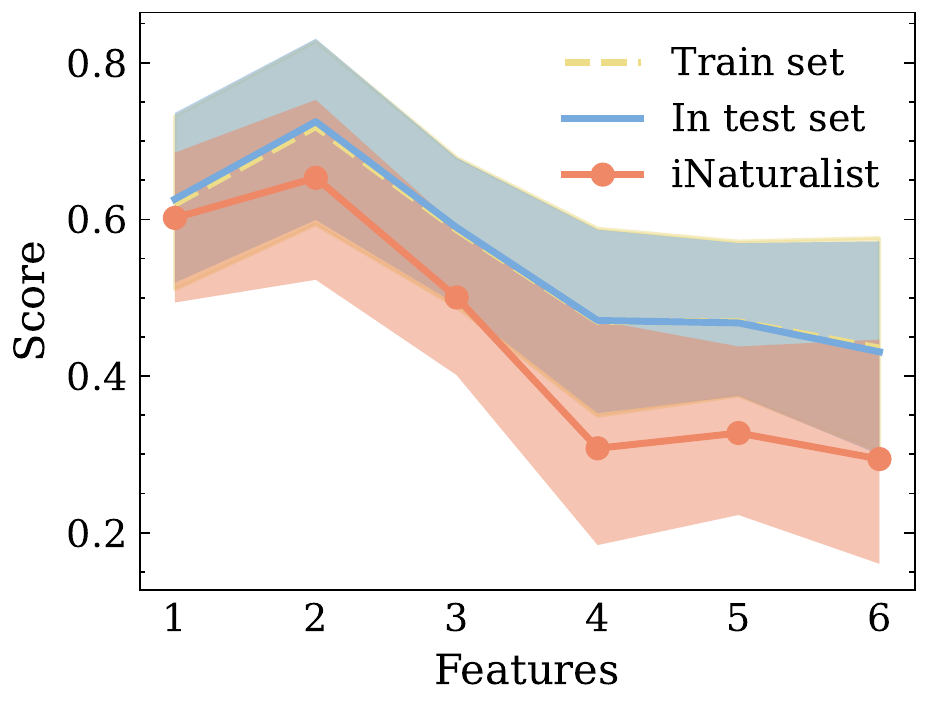}
        \caption{Trajectories for iNaturalist.}
        \label{fig:dn_time_series_inaturalist_appendix}
    \end{subfigure}
    \hfill
    \begin{subfigure}[b]{0.329\textwidth}
        \centering
        \includegraphics[width=\textwidth]{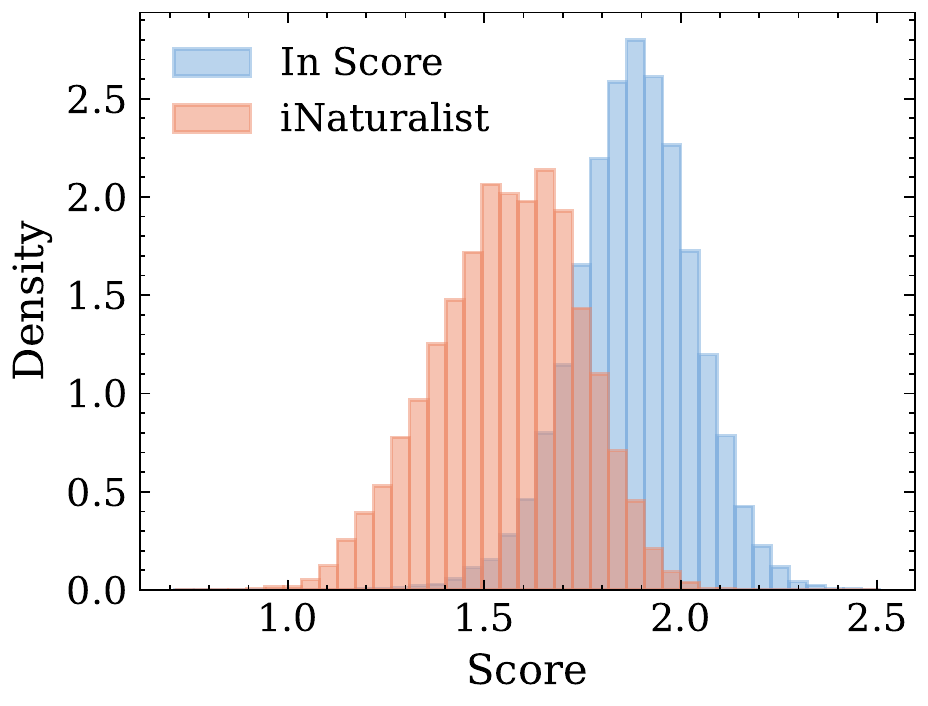}
        \caption{Score histogram for iNaturalist.}
        \label{fig:dn_histogram_inaturalist_appendix}
    \end{subfigure}
    \hfill
    \begin{subfigure}[b]{0.329\textwidth}
        \centering
        \includegraphics[width=\textwidth]{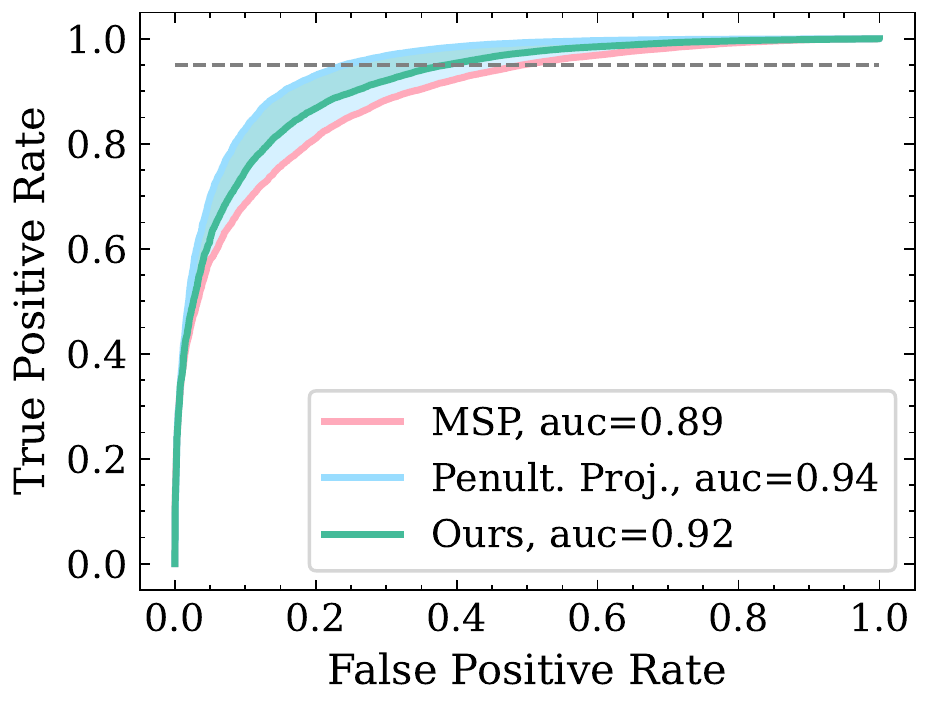}
        \caption{ROC curve for iNaturalist.}
        \label{fig:dn_roc_inaturalist_appendix}
    \end{subfigure}
    \hfill

    \begin{subfigure}[b]{0.329\textwidth}
        \includegraphics[width=\textwidth]{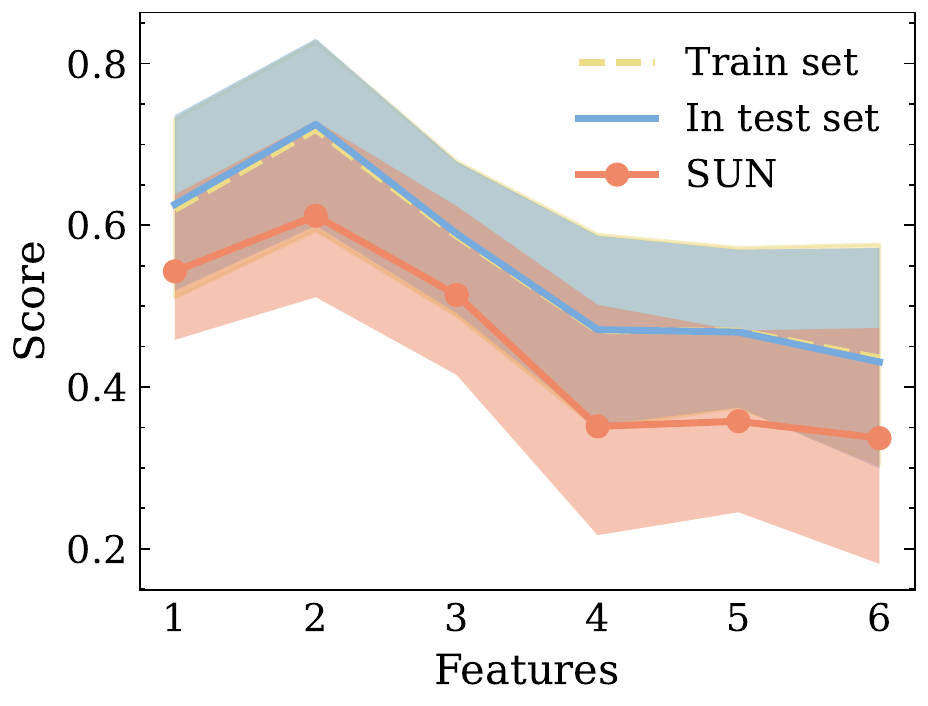}
        \caption{Trajectories for SUN.}
        \label{fig:dn_time_series_sun_appendix}
    \end{subfigure}
    \hfill
    \begin{subfigure}[b]{0.329\textwidth}
        \centering
        \includegraphics[width=\textwidth]{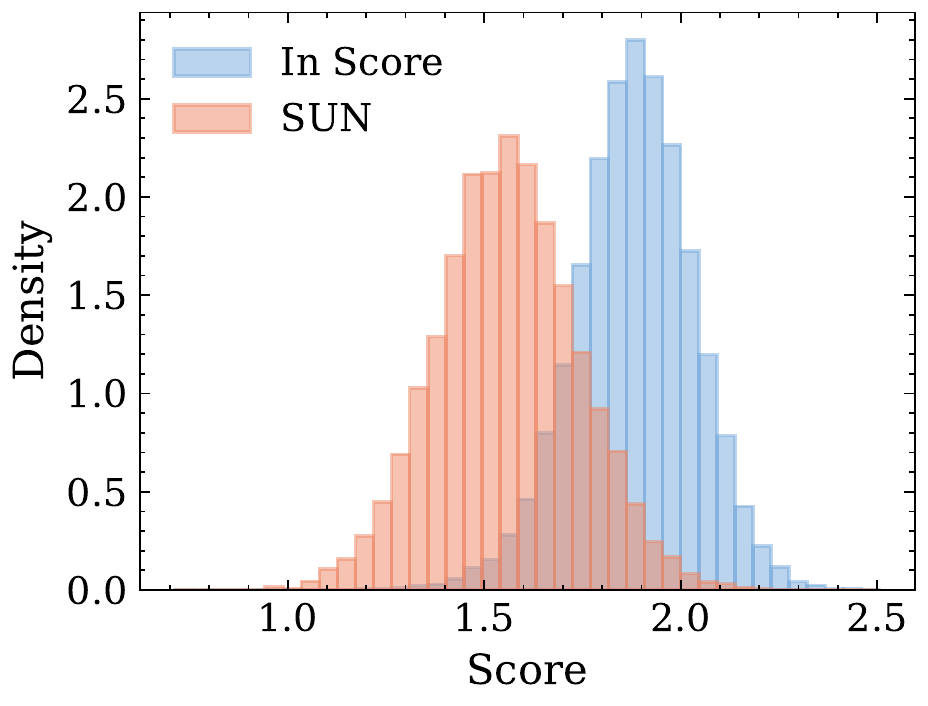}
        \caption{Score histogram for SUN.}
        \label{fig:dn_histogram_sun_appendix}
    \end{subfigure}
    \hfill
    \begin{subfigure}[b]{0.329\textwidth}
        \centering
        \includegraphics[width=\textwidth]{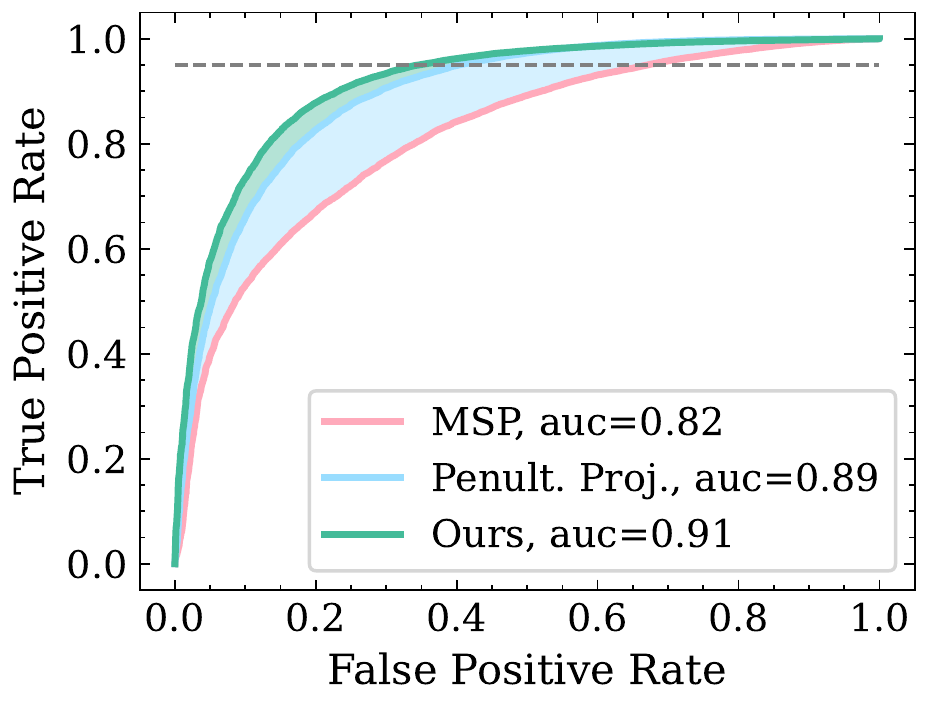}
        \caption{ROC curve for SUN.}
        \label{fig:dn_roc_sun_appendix}
    \end{subfigure}
    \hfill
     
    \begin{subfigure}[b]{0.329\textwidth}
        \includegraphics[width=\textwidth]{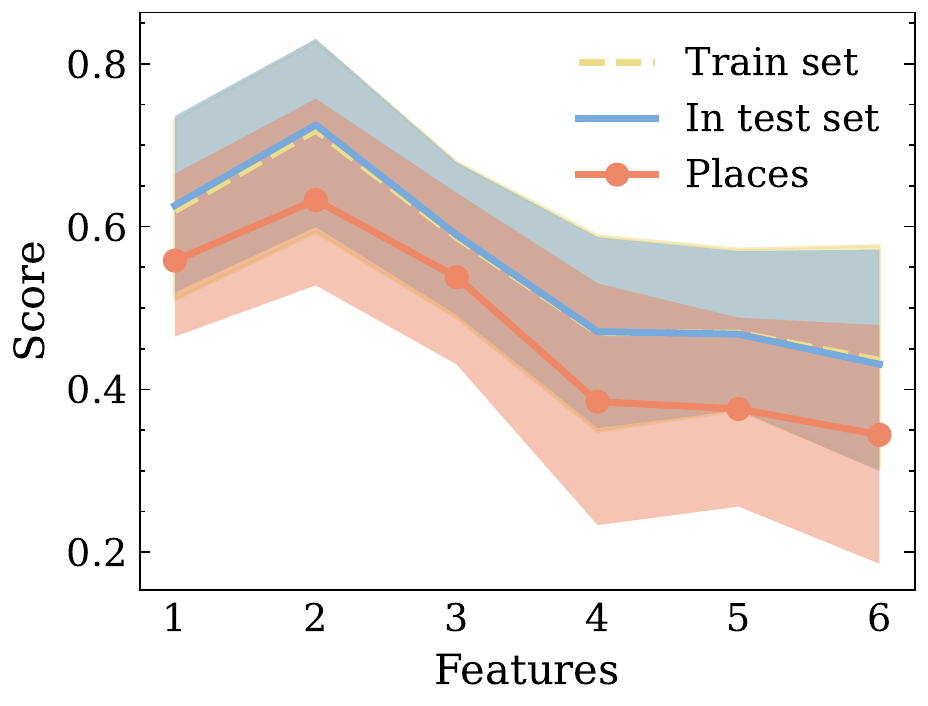}
        \caption{Trajectories for Places.}
        \label{fig:dn_time_series_places_appendix}
    \end{subfigure}
    \hfill
    \begin{subfigure}[b]{0.329\textwidth}
        \centering
        \includegraphics[width=\textwidth]{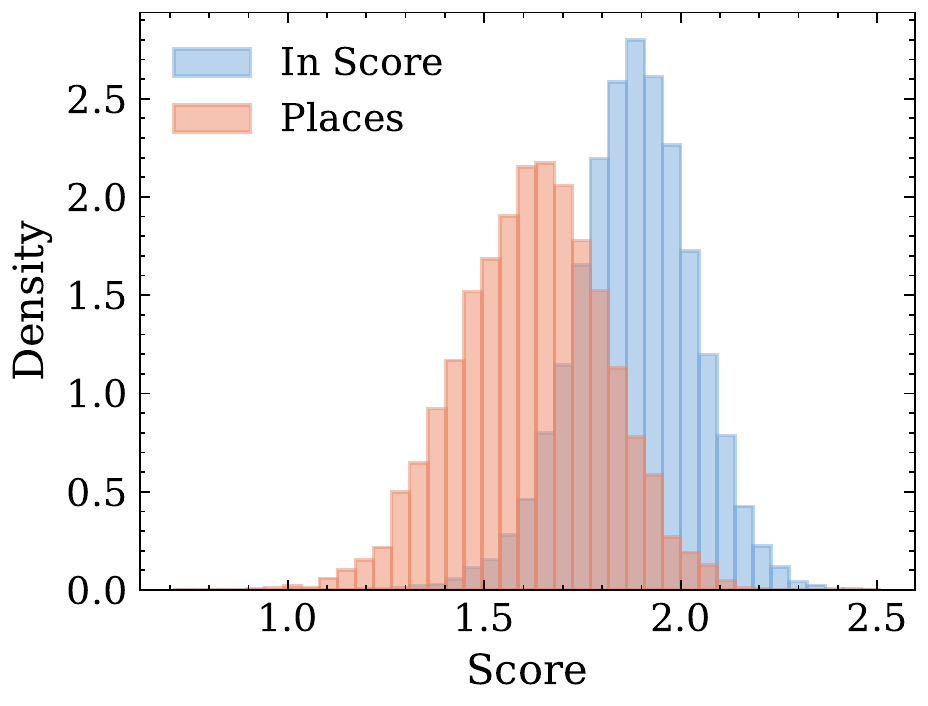}
        \caption{Score histogram for Places.}
        \label{fig:dn_histogram_places_appendix}
    \end{subfigure}
    \hfill
    \begin{subfigure}[b]{0.329\textwidth}
        \centering
        \includegraphics[width=\textwidth]{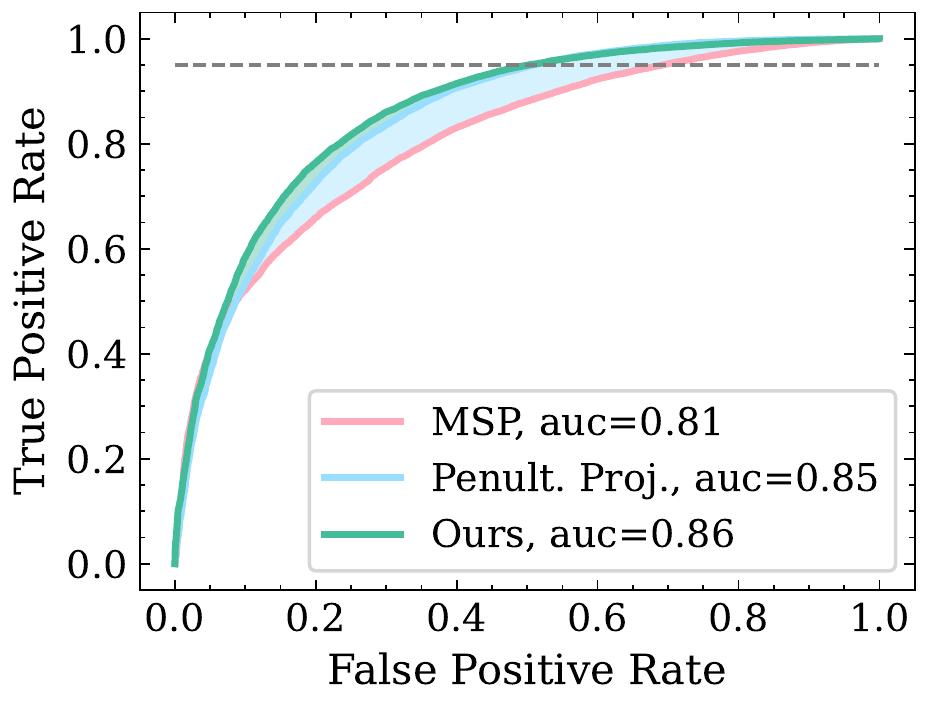}
        \caption{ROC curve for Places.}
        \label{fig:dn_roc_places_appendix}
    \end{subfigure}
    \hfill
    
    \begin{subfigure}[b]{0.329\textwidth}
        \includegraphics[width=\textwidth]{images/densenet121_ilsvrc2012/textures/time_series_L2_COSINE_pred.pdf}
        \caption{Trajectories for Textures.}
        \label{fig:dn_time_series_textures_appendix}
    \end{subfigure}
    \hfill
    \begin{subfigure}[b]{0.329\textwidth}
        \centering
        \includegraphics[width=\textwidth]{images/densenet121_ilsvrc2012/textures/hist_L2_COSINE_pred.pdf}
        \caption{Score histogram for Textures.}
        \label{fig:dn_histogram_textures_appendix}
    \end{subfigure}
    \hfill
    \begin{subfigure}[b]{0.329\textwidth}
        \centering
        \includegraphics[width=\textwidth]{images/densenet121_ilsvrc2012/textures/roc_L2_COSINE_pred.pdf}
        \caption{ROC curve for Textures.}
        \label{fig:dn_roc_textures_appendix}
    \end{subfigure}
    \hfill
       \caption{Average trajectories, OOD detection score histogram and ROC curve for the DenseNet-121 model.}
    \label{fig:dn_roc_hist_appendix}
\end{figure}

We display additional plots in Figure~\ref{fig:dn_roc_hist_appendix} for the observed functional data, the histogram of our scores showing separation between in-distribution and OOD data and the ROC curves for the DenseNet-121 model, as an illustrative example. 

\end{document}